\documentclass[letterpaper]{article} % DO NOT CHANGE THIS
\usepackage{aaai25}  % DO NOT CHANGE THIS
\usepackage{times}  % DO NOT CHANGE THIS
\usepackage{helvet}  % DO NOT CHANGE THIS
\usepackage{courier}  % DO NOT CHANGE THIS
\usepackage[hyphens]{url}  % DO NOT CHANGE THIS
\usepackage{graphicx} % DO NOT CHANGE THIS
\usepackage{natbib}  % DO NOT CHANGE THIS AND DO NOT ADD ANY OPTIONS TO IT
\usepackage{caption} % DO NOT CHANGE THIS AND DO NOT ADD ANY OPTIONS TO IT
\usepackage{algorithm}
\usepackage{algorithmic}
\usepackage{graphicx}
\usepackage{amsmath}
\usepackage{amssymb}
\usepackage{xcolor}
\usepackage{tikz}
\usepackage{comment}
\newcommand{\defi}{\overset{\mathrm{def}}{=}}
\let\vec=\overrightarrow
\newtheorem{theorem}{Theorem}

\newtheorem{definition}{Definition}
\newtheorem{proposition}{Proposition}

\definecolor{CBRed}{RGB}{0,0,0}
\colorlet{red}{CBRed}

\usepackage{newfloat}
\usepackage{listings}
\DeclareCaptionStyle{ruled}{labelfont=normalfont,labelsep=colon,strut=off} % DO NOT CHANGE THIS
\floatstyle{ruled}
\newfloat{listing}{tb}{lst}{}
\floatname{listing}{Listing}
\title{Reasoning about Actual Causes in Nondeterministic Domains -- Extended Version}
\author{
    Shakil M.\ Khan\textsuperscript{\rm 1},
    Yves Lesp\'{e}rance\textsuperscript{\rm 2},
    Maryam Rostamigiv\textsuperscript{\rm 1}
}
\affiliations{
    \textsuperscript{\rm 1}University of Regina, Regina, SK, Canada\\
    \textsuperscript{\rm 2}York University, Toronto, ON, Canada\\
    shakil.khan@uregina.ca, lesperan@eecs.yorku.ca, maryam.rostamigiv@uregina.ca
}

\begin{document}

\maketitle

\begin{abstract}
        Reasoning about the causes behind observations is crucial to the formalization of rationality. While extensive research has been conducted on root cause analysis, most studies have predominantly focused on deterministic settings. In this paper, we investigate causation in more realistic nondeterministic domains, where the agent does not have any control on and 
        may not know %YL changed
        %does not know 
        the choices that are made by the environment. We build on recent preliminary work on actual causation in the nondeterministic situation calculus to formalize more sophisticated forms of reasoning about actual causes in such domains.
        We investigate the notions of “Certainly Causes” and “Possibly Causes” that enable the representation of actual cause 
        for agent actions %YL added
in these domains. We then show how regression in the situation calculus can be extended to reason about such notions of actual causes.
\end{abstract}

% Uncomment the following to li33nk to your code, datasets, an extended version or similar.

\section{Introduction}
%This work introduces a new way of looking at causality in nondeterministic (ND) systems that do not follow certain patterns \textbf{[YL: ??]}. When one agent action is influenced by environmental reaction can be led to many different outcomes, it is essential to create a logic that can pinpoint and examine actual causes. \textbf{[YL: Clarify]} This challenge is especially important in fields like AI, robotics, and autonomous systems, where decisions must consider uncertainty and change.
%
The term \emph{causation} refers to a collection of closely-related important philosophical problems dealing with causes and their effects that has been studied since the time of Aristotle. Determining the ``actual'' causes of an observed effect, which are events chosen from a recorded history of actions that occurred prior to the observation of the effect (also known as the scenario) is one such problem (called ``efficient'' cause in Aristotelian lingo) that has been extensively researched. Motivated by David Hume's philosophical work and Herbert Simon's early contributions, Pearl \cite{Pearl98,Pearl00}, and later Halpern \cite{Halpern00,Halpern15,Halpern16}, Halpern and Pearl \cite{HalpernP05}, and others \cite{EiterL02,Hopkins05,HopkinsP07} developed computational formalizations of this problem within Structural Equations Models (SEM). While their inspirational work significantly advanced this field, their approach based on SEM has been nevertheless criticized for its limited expressiveness \cite{Hopkins05,HopkinsP07,GlymourDGERSSTZ10}, {\color{CBRed} and others have expanded SEM with additional features \cite{Leitner-FischerL13} or proposed alternate formalizations of actual cause, e.g.\ \cite{Bochman18,BeckersV18,LimaL24}.
}
\par
In response to these criticisms, in recent years researchers have become increasingly interested in studying causation within more expressive action-theoretic frameworks, in particular in that of the situation calculus \cite{BatusovS17,BatusovS18,KhanS20}. Among other things, this allows one to formalize causation from the perspective of individual agents by defining a notion of epistemic causation \cite{KL21} and by supporting causal reasoning about conative effects, which in turn has proven useful for explaining agent behaviour using causal analysis \cite{KR23} and has the potential for defining important concepts such as responsibility and blame \cite{YazdanpanahGSDJNR23}. 
\par%777
While there has been a lot of work on actual causation, 
%including proposals in formal action-theoretic frameworks, 
the vast majority of the work in this area has focused on deterministic domains. 
However, a distinguishing feature of the real world is that change is often unpredictable. Very few studies address causation in nondeterministic systems, and those that do, are formalized in SEM-based causal models that are known to have limited expressiveness and suffer from a variety of problems. For instance, recently Beckers \shortcite{SB24} presented an extension of causal models to deal with nondeterminism. However, despite improving on expressivity of causal models, it is not clear how one can formalize various aspects of action-theoretic/dynamic frameworks there, e.g.\ non-persistent change supported by fluents, possible dependency between events, temporal order of event occurrence, etc.
\par
%%{\color{blue}YL updated: 
To deal with this, building on previous work on actual causation in the situation calculus \cite{BatusovS18,KL21}, more recently \cite{RKLY24} formalized actual causes in more expressive nondeterministic action-theoretic domains. They used the nondeterministic situation calculus \cite{DeGiacomoLespKR21} as their base framework. They introduced notions of “Certainly Causes” and “Possibly Causes” that enable the representation of actual cause when the agent does not control her actions' outcomes and does not know the choices that are made by the environment. However, this early work formalizes reasoning about causes in these domains by considering all possible evolutions of the scenario, which makes reasoning computationally intractable.\footnote{They reason about “Certainly Causes” by considering all the 
possible executions of the scenario, 
%i.e., all the executions of the sequence of agent actions
and determining whether the given action was an actual cause in each of them. However, the number of distinct executions of such a scenario/agent action sequence  is exponential in the number of nondeterministic actions in the scenario.} In this paper, we build on this preliminary work and formalize more effective forms of reasoning about actual causes in such nondeterministic domains.  In particular, we extend regression in the situation calculus to provide a more effective mechanism to reason about actual causes.
\par
The paper is structured as follows. In the next section, we provide an overview of the situation calculus and nondeterministic situation calculus (NDSC) and introduce our running example. Then, we examine the definition of actual cause proposed earlier. After that, we present the definitions of various causal notions in the NDSC. Next, we formalize reasoning about %the aforementioned notions of 
actual causes in the NDSC. In particular, we prove some properties showing how regression in the situation calculus can be extended to deal with these notions. 
%we demonstrate how causes can be simplified into regressable formulae. 
Finally, we conclude the paper with some discussion. %in Section 6.

\section{Preliminaries}
%=============================
%=============================
%*****************************
%=============================
%=============================
\paragraph{\textup{\textbf{Situation Calculus (SC).}}}
The situation calculus is a well-known second-order language for representing and reasoning about dynamic worlds \cite{McCarthyH69,Reiter01}. In the SC, all changes %in the world 
are due to named actions, which are terms in the language. Situations represent a possible world history resulting from performing some actions. The constant $S_0$ is used to denote the initial situation where no action has been performed yet. The distinguished binary function symbol $\mathit{do}(a, s)$ denotes the successor situation to $s$ resulting from performing the action $a$. The expression $\mathit{do}([a_1,\cdots,a_n],s)$ represents the situation resulting from executing actions $a_1,\cdots,a_n$, starting with situation $s$. As usual, a relational/functional fluent representing a property whose value may change from situation to situation takes a situation term as its last argument. There is a special predicate $\mathit{Poss}(a,s)$ used to state that action $a$ is executable in situation $s$. Also, the special binary predicate $s \sqsubset s'$ represents that $s'$ can be reached from situation $s$ by executing some sequence of actions.  $s\sqsubseteq s'$ is an abbreviation of $s\sqsubset s'\lor s=s'$. $s < s'$ is an abbreviation of $s\sqsubset s'\land\mathit{Executable}(s')$, where $\mathit{Executable}(s)$ is defined as $\forall a',s'.\;do(a',s')\sqsubseteq s \supset\mathit{Poss}(a', s')$, i.e.\ every action performed in reaching situation $s$ was possible in the situation in which it occurred. $s \leq s'$ is an abbreviation of $s<s'\vee s=s'$.
\par
In the SC, a dynamic domain is specified using a basic action theory (BAT) $\mathcal{D}$ that includes the following sets of axioms: (i) (first-order or FO) initial state axioms $\mathcal{D}_{S_0}$, which indicate what was true initially; (ii) (FO) action precondition axioms $\mathcal{D}_\mathit{ap}$, characterizing $\mathit{Poss}(a, s)$; (iii) (FO) successor-state axioms $\mathcal{D}_\mathit{ss}$, 
specifying how the fluents change when an action is performed and providing a solution to the frame problem;
%indicating precisely how the fluents change;
(iv) (FO) unique-names axioms $\mathcal{D}_{\mathit{una}}$ for actions, stating that different action terms represent distinct actions; and (v) (second-order or SO) domain-independent foundational axioms $\Sigma$, describing the structure of situations \cite{LevPirRei98}.
\par
%{\color{red} 
A key feature of BATs is the existence of a sound and complete {\em regression mechanism} for answering queries about situations resulting from performing a sequence of actions \cite{PirRei99, Reiter01}. In a nutshell, the regression operator $\mathcal{R}^*$ reduces a formula $\phi$ about a particular future situation to an equivalent formula $\mathcal{R}^*[\phi]$ about the initial situation $S_0$. 
A formula $\phi$ is regressable if and only if (i) all situation terms in it are of the form $do([a_1, \dots,a_n],S_0)$, (ii) in every atom of the form $\mathit{Poss}(a,\sigma)$, the action function is specified, i.e., $a$ is of the form $A(t_1, \dots, t_n)$, (iii) it does not quantify over situations, and (iv) it does not contain $\sqsubset$ or equality over situation terms. Thus in essence, a formula is regressable if it does not contain situation variables. 
\par
%The regression operator transforms the regressable formula $\varphi$ into an equivalent formula that needs to hold in the initial situation for $\varphi$ to be true after a sequence of actions. 
In the following, we define a one-step variant of $\mathcal{R}^*,$ $\mathcal{R}$.
%restricted to relational fluents (for simplicity). 
%
\begin{definition}[The Regression Operator]\label{regressionDef}\mbox{}\\
(1) When $\phi$ is a non-fluent atom, including equality atoms without functional fluents as
arguments, or when $\phi$ is a fluent atom, whose situation argument is $S_0$, $\mathcal{R}[\phi]=\phi$.\\
(2a) %There are three cases, (a) for ordinary non-functional fluents, (b) for equality with at least one functional fluent, and (c) for the regression of $\mathit{Poss}$ literals. 
For a non-functional fluent $F$, whose successor-state axiom in $\mathcal{D}$ is  $F(\vec{x},do(a,s))\equiv\Phi_F(\vec{x},a,s)$, $\mathcal{R}[F(\vec{t},do(\alpha,\sigma))]=\Phi_F(\vec{t},\alpha,\sigma)$. %In other words, the fluent $F(\vec{t}, do(\alpha, \sigma))$ is replaced by the appropriate instance of the right-hand side of the equivalence in F's successor state axiom.
\\
(2b) For an equality literal with a functional fluent $f$, whose successor-state axiom is $f(\vec{x},do(a,s))=y\equiv\Phi_f(\vec{x},y,a,s),$ $\mathcal{R}[f(\vec{t},do(\alpha,\sigma))=t']=\Phi_f(\vec{t},t',\alpha,\sigma).$ 
\\
(2c) For a $\mathit{Poss}$ literal with the action precondition
axiom of the form $\mathit{Poss}(A(\vec{x}),s)\equiv\Pi_{A}(\vec{x},s)$, $\mathcal{R}[\mathit{Poss}(A(\vec{t}),\sigma)]\equiv\mathcal{R}[\Pi_{A}(\vec{t},\sigma)]$.\\
(3) For any non-atomic formulae, regression is defined inductively: $\mathcal{R}[\neg \phi]=\neg \mathcal{R}[\phi]$, $\mathcal{R}[\phi_1\wedge \phi_2]=\mathcal{R}[\phi_1]\wedge\mathcal{R}[\phi_2]$, $\mathcal{R}[\exists v.\;\phi]=\exists v.\;\mathcal{R}[\phi]$.
\end{definition}
$\mathcal{R}^*$ can then be defined as the repeated application of $\mathcal{R}$ until further applications leave the formula unchanged.
%%}
%
\par
Another key result about BATs is the relative satisfiability theorem \cite{PirRei99, Reiter01}: $\mathcal{D}$ is satisfiable if and only if $\mathcal{D}_{S_0}\cup \mathcal{D}_\mathit{una}$ is satisfiable (the latter being a purely first-order theory).
\paragraph{\textup{\textbf{Nondeterministic Situation Calculus (NDSC).}}}
An important limitation of the standard SC and BATs is that atomic actions are deterministic. De Giacomo and Lesp\'{e}rance (DL21) \cite{DeGiacomoLespKR21} proposed a simple extension of the framework to handle nondeterministic actions while preserving the solution to the frame problem.
For any primitive action by the agent in a nondeterministic domain, there can be a number of different outcomes. (DL21) takes the outcome as being determined by the agent's action and the environment's reaction to this action. This is modeled by having every action type/function $A(\vec{x}, e)$ take an additional environment reaction parameter $e$, ranging over
a new sort \emph{Reaction} of environment reactions. The agent cannot control the environment reaction, so it performs the reaction-suppressed version of the action $A(\vec{x})$ and the environment then selects a reaction $e$ to produce the complete action $A(\vec{x}, e)$. 
We call the reaction-suppressed version of the action $A(\vec{x})$ an \emph{agent action} and the full version of the action $A(\vec{x}, e)$ a \emph{system action}.

We represent nondeterministic domains using action theories called Nondeterministic Basic Action Theories (NDBATs), which can be seen as a special kind of BAT, where $(1)$ every agent action takes an environment reaction parameter; $(2)$ for each agent action we have an agent action precondition formula, $\mathit{Poss}_{ag}(a(\vec{x}), s)\defi\phi_{a}^\mathit{agPoss} (\vec{x}, s)$; $(3)$ for each agent action we have a reaction independence requirement, stating that the precondition for the agent action is independent of any environment reaction $\forall e.\;\mathit{Poss}(a(\vec{x}, e), s)\supset\mathit{Poss}_\mathit{ag}(a(\vec{x}), s)$; $(4)$ for each agent action we also have a reaction existence requirement, stating that if the precondition of the agent action holds then there exists a reaction to it which makes the complete system action executable, i.e., the environment cannot prevent the agent from performing an action when its agent action precondition holds $\mathit{Poss}_\mathit{ag}(a(\vec{x}),s)\supset\exists e.\;\mathit{Poss}_\mathit{ag}(a(\Vec{x},e), s)$.
The above requirements \emph{must} be entailed by the action theory for it to be an NDBAT.

A NDBAT $\mathcal{D}$ is the union of the following disjoint sets: including (1) foundational axioms, (2) unique-names axioms for actions, (3) axioms describing the initial situation, (4) successor-state axioms indicating how fluents change after system actions, %$Poss(a(\Vec{x}, e), s) \equiv \phi^{Poss}_{a}(\Vec{x}, e, s)$ 
and (5) system action precondition axioms, indicating when system actions can occur; while these axioms generally follow the form: $\mathit{Poss}(a(\vec{x}, e), s)\equiv\phi^\mathit{Poss}_{a} (\vec{x}, e, s)$, in practice, these axioms often take the form: $\mathit{Poss}(a(\vec{x}, e), s)\equiv\mathit{Poss}_{ag}(a(\vec{x}), s)\wedge\varphi^\mathit{Poss}(\vec{x}, e, s)$, where $\mathit{Poss}_{ag}(a(\vec{x}), s)$ denotes conditions necessary for the agent action $a(\vec{x})$ to occur and $\phi^\mathit{Poss}_{a} (\vec{x}, e, s)$ captures additional conditions influenced by the environment's response.

%\subsection*{The Situation Calculus}
%The SC is recognized a predicate logic language for representing and reasoning about dynamically changing worlds \cite{McCarthy1969:AI} and \cite{Reiter01}. In this framework, any change that happens in the world is because of actions. These actions are expressed using specific terms in the language. A sequence of actions and their outcomes over time is depicted using a term called a "situation". We designate the initial situation as $S_0$. To show a sequence of actions leading to different situations, we use the function symbol $do$ such as $do(a, s)$ denotes the successor situation resulting from performing action a in situation $s$. Certain properties or characteristics of the world that change from one situation to another are called "fluents". They're represented by symbols, which include a situation term as part of their description. In this framework, we can describe a dynamic environment using a Basic Action Theory (BAT). This theory includes successor state axioms, which define how the world changes due to actions, and it addresses challenges like the frame problem. To determine whether an action is possible in a situation, we use precondition axioms. When we say $\mathit{Executable}(s)$, it means that every action that led to situation $s$ was indeed possible in the situations where they occurred.

%\paragraph{\textup{\textbf{Executability and Projection.}}}
\paragraph{\textup{\textbf{Projection and Executability.}}}
In the NDSC, executing an agent action in a situation may result in different situations and outcomes depending on the environment reaction.  
To capture this, (DL21) introduced the defined predicate $Do_{ag}(\vec{a},s,s')$, meaning that the system may reach situation $s'$
when the agent executes the sequence of agent actions $\vec{a}$ depending on environment reactions:
%\begin{small}
\[\begin{array}{l} 
\hspace{-.05 cm}Do_{ag}(\epsilon,s,s') \defi s=s',
\\\hspace{4 mm}\mbox{where $\epsilon$ is the empty sequence of actions},
\\
\hspace{-.05 cm}Do_{ag}([A(\vec{x}),\sigma],s,s') \defi \\
   \hspace{.5 cm} \exists e.\;\mathit{Poss}(A(\vec{x},e),s) \land Do_{ag}(\sigma,do(A(\vec{x},e),s),s').
\end{array} \]
%\end{small}
A condition $\phi$ may hold after some executions of a sequence of agent actions $\vec{a}$ starting in situation $s$, %i.e., possibly after $\vec{a}$ or 
denoted by $\mathit{PAfter}(\vec{a},\phi, s)$, or it may hold after all executions of $\vec{a}$ in $s$, %i.e., certainly after $\vec{a}$ or 
denoted by $\mathit{CAfter}(\vec{a},\phi, s)$. Formally: %These notions are defined as follows:
%\begin{small}
\[\begin{array}{l}
\mathit{PAfter}(\vec{a}, \phi, s) \defi \exists s'. Do_{ag}(\vec{a}, s, s') \wedge \phi[s'],
\\
\mathit{CAfter}(\vec{a}, \phi, s) \defi \forall s'. Do_{ag}(\vec{a}, s, s') \supset \phi[s'].
\end{array}\]
%\end{small}
%
%
%
Two different notions of executability of %an agent action sequence 
$\vec{a}$ are also defined (see \cite{DeGiacomoLespKR21}).
\begin{comment}
An agent action sequence $\vec{a}$ can be executable in situation $s$ under all possible environment reactions, i.e., be certainly executable, denoted by $\mathit{CertainlyExecut}\-\mathit{able}(\vec{a}, s)$. Alternatively, it can be executable in situation $s$ under some environment reactions, i.e., be possibly executable, or $\mathit{PossiblyExecutable}(\vec{a}, s)$.\\
%\textcolor{red}{MR: I think it would be better to replace $a$ by $s$ in $CertainlyExecutable(\vec{a},a)$ and $PossiblyExecutable(\vec{a},a)$.}
%\begin{small}
\[\begin{array}{l}
    \mathit{CertainlyExecutable}(\epsilon,s) \defi \mathit{True},\\
    \mathit{CertainlyExecutable}([A(\vec{x}), \sigma], s) \defi 
    \mathit{Poss}_\mathit{ag}(A(\vec{x}),s) \land \\
   \hspace{4.7cm} \forall s'. Do_{ag}(A(\vec{x}), s, s') \supset
    \mathit{CertainlyExecutable}(\sigma, s').\\[.5ex]
   \mathit{PossiblyExecutable}(\epsilon,s) \defi \mathit{True},\\
   \mathit{PossiblyExecutable}([A(\vec{x}), \sigma], s) \defi
        \exists s'. Do_{ag}(A(\vec{x}), s, s') \land
   \mathit{PossiblyExecutable}(\sigma, s').
  \end{array} \]
%\end{small}
\end{comment}

%\example\label{example1}
\paragraph{\textup{\textbf{Example.}}} Our running example involves a robot navigating between different locations and communicating.
We take communication to be subject to interference and assume that the robot can communicate at a given location if the location is not risky and it has not become vulnerable.
%Some of the locations are identified as high-risk areas.
The robot can move between locations if they are connected and communicate from current location (represented using agent actions $\mathit{move}(i,j)$ and $\mathit{comm}(i)$). While moving to a location, the robot faces the possibility of becoming vulnerable if that location is a risky one.
% and the robot can communicate at a given location if the location is not risky and he has not become vulnerable.
Thus the agent action $\mathit{move}(i,j)$ is associated with the system action $\mathit{move}(i,j,e)$, where the environment reaction $e$ can be either $\mathit{Vul}$ (for becoming vulnerable) or $\mathit{NotVul}$ (for not becoming vulnerable). The communicate agent action on the other hand has only one successful environment reaction and so it is associated with system action $\mathit{comm}(i,e),$ where $e=\mathit{Succ}$. 
\par
The precondition axioms for these agent and system actions are as follows.\par\vspace{-4 mm}
\begin{small}
\begin{eqnarray*}
&&\hspace{-5 mm}(1)\;\mathit{Poss}_\mathit{ag}(\mathit{move}(i,j),s)\equiv\mathit{At}(i,s)\land \mathit{Connected}(i,j),\\
%\vee\mathit{Connected}(j,i)),\\
&&\hspace{-5 mm}(2)\;\mathit{Poss}_\mathit{ag}(\mathit{comm}(i),s)\equiv\neg\mathit{Vul}(s)\wedge\neg\mathit{Risky}(i,s),\\
&&\hspace{-5 mm}(1')\;\mathit{Poss}(\mathit{move}(i,j,e),s)\equiv\mathit{Poss}_\mathit{ag}(move(i,j),s)\\
&&\hspace{2 mm}\mbox{}\land(\mathit{Risky}(j,s)\supset(e=\mathit{Vul} \vee e=\mathit{NotVul}))\\
&&\hspace{2 mm}{}\land
(\neg\mathit{Risky}(j,s)\supset e=\mathit{NotVul}),\\
%\end{eqnarray*}
%\begin{eqnarray*}
&&\hspace{-5 mm}(2')\;\mathit{Poss}(\mathit{comm}(i, e),s)\equiv\mathit{Poss}_\mathit{ag}(\mathit{comm}(i),s) \wedge e=\mathit{Succ}.
\end{eqnarray*}
\end{small}
\par
The fluents in this example are $\mathit{Vul}(s)$, which denotes that the robot is vulnerable in situation $s$, $\mathit{At}(i,s)$, which states that the robot is at location $i$ in $s$, and $\mathit{Risky}(i,s)$, which indicates that the location $i$ is risky in $s$. Certain locations are risky initially and they remain the only risky ones when actions are performed. Also, we have a non-fluent $\mathit{Connected}(i, j)$ to indicate that there is an edge from $i$ to $j$.
\par
The successor-state axioms (SSA) for these are:\par\vspace{-4 mm}
\begin{small}
\begin{eqnarray*}
&&\hspace{-7 mm}(3)\;\mathit{At}(j,do(a,s))\equiv\exists i,e.\; a=\mathit{move}(i,j,e)\\
&& \hspace{15 mm}\mbox{}\lor(\mathit{At}(j,s)\wedge\forall j',e'.\;\neg(a=\mathit{move}(j,j',e')),\\
&&\hspace{-7 mm}(4)\;\mathit{Vul}(do(a,s))\equiv\exists i,j.\;a=\mathit{move}(i,j,\mathit{Vul}) \vee \mathit{Vul}(s),\\
&&\hspace{-7 mm}(5)\;\mathit{Risky}(i,do(a,s))\equiv\mathit{Risky}(i,s).
\end{eqnarray*}
\end{small}
\par
We also have the following initial state axioms:\par\vspace{-4 mm}
\begin{small}
\begin{eqnarray*}
&&\hspace{-8 mm}(6)\neg\mathit{Vul}(S_0),\; 
(7)\mathit{At}(I_0,S_0),\; 
%&&(7)\;\mathit{Dest}(I, S_0)\equiv I=I_3$,\\ 
(8)\mathit{Risky}(i,S_0)\equiv i=I_1\vee i=I_2.
\end{eqnarray*}
\end{small}
%\textcolor{red}{MR:I think both $I_1$ and $I_2$ are risky locations.}
\par
\begin{figure}[t]
\begin{center}
\begin{tikzpicture}
    \node (v1) at (-2,1) {};
    \node (v2) at (-1,1) {};
    \node (v3) at (0,1) {};
    \node (v4) at (1,1) {};
    
    \draw (v1) edge (v2);
    \draw (v2) edge (v3);
    \draw (v3) edge (v4);
    
    % Adding small circles at connection points
    \node at (-2,1) [circle, fill, inner sep=0.5mm] {};
    \node at (-1,1) [circle, fill, inner sep=0.5mm] {};
    \node at (0,1) [circle, fill, inner sep=0.5mm] {};
    \node at (1,1) [circle, fill, inner sep=0.5mm] {};
    
    % Adding labels above the small circles
    \node[above] at (-2, 1) {$I_0$};
    \node[above] at (-1,1) {$I_1$};
       \node[above] at (-1,.6) {\tiny{$\mathit{Risky}$}};
    \node[above] at (0,1) {$I_2$};
    \node[above] at (0,.6) {\tiny{$\mathit{Risky}$}};
    \node[above] at (1,1) {$I_3$};
\end{tikzpicture}
\end{center}
\mbox{}\vspace{-6 mm}
\caption{Interconnections between locations}\label{fig:interconnect}
\end{figure}
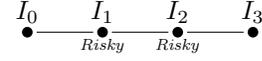

Finally, there are $4$ locations, $I_0$ to $I_3$, and the interconnections between these
axiomatized using $\mathit{Connected}$ is depicted in Fig.~\ref{fig:interconnect}.
%Finally, there are $4$ locations $I_0$ to $I_3$ in this domain, and the interconnections between these locations are given by the following axiom (see Fig.~\ref{fig:interconnect}): %\textbf{[YL: add figure?]}.
%\[\begin{array}{l}(9)\;\mathit{Connected}(i,j)\equiv (i=I_0\land j=I_1)\lor(i=I_1\land j=I_2)\lor(i=I_2\land j=I_3) \lor {}\\
%\hspace{10em}
%(i=I_1\land j=I_0)\lor(i=I_2\land j=I_1)\lor(i=I_3\land j=I_2).\end{array}\]
We will call this NDBAT $\mathcal{D}_1$.
\section{Actual Achievement Cause in the SC}
%=============================
%=============================
%*****************************
%=============================
%=============================
%\section{Actual Achievement Cause in the SC}
Given a history of actions/events (often called a scenario) and an observed effect, \emph{actual causation} involves figuring out which of these actions are responsible for bringing about this effect.\footnote{We use actions and events interchangeably.} When the effect is assumed to be false before the execution of the actions in the scenario and true afterwards, the notion is referred to as \emph{achievement (actual) causation}. Based on Batusov and Soutchanski's \shortcite{BatusovS18} original proposal, Khan and Lesp\'{e}rance \shortcite{KL21} (KL21) recently offered a definition of achievement cause in the SC. Both of these frameworks assume that the scenario is a linear sequence of actions, i.e.\ no concurrent actions are allowed. (KL21)'s proposal can deal with epistemic causes and effects; e.g., an agent may analyze the cause of some newly acquired knowledge, and the cause may include some knowledge-producing action, e.g.\ $\mathit{inform}$. They showed that an agent may or may not know all the causes of an effect, and can even know some causes while not being sure about others. 
\par 
%In this framework, causes are computed relative to a {\em causal setting} consisting of a domain theory $\mathcal{D}$, a scenario $\sigma$ (which, again, is a linear sequence of actions), and an effect $\varphi$. Since all changes in the SC result from actions, they identified the potential causes with a set of ground action terms occurring in $\sigma$. However, since $\sigma$ might include multiple occurrences of the same action, one also needs to identify the situations where these actions were executed; for brevity, we will ignore this component here. The underlying idea of computing causes is as follows. If some action $\alpha$ of the action sequence in $\sigma$ triggers the formula $\varphi$ to change its truth value from false to true relative to $\mathcal{D}$, and if there are no actions in $\sigma$ after $\alpha$ that change the value of $\varphi$ back to false, then $\alpha$ is an actual cause of achieving $\varphi$ in $\sigma$. Moreover, note that $\alpha$ might have been non-executable initially; so other preceding actions that contributed to ensuring that its preconditions are brought about must also be considered as (indirect) cause of $\varphi$. Similarly, $\alpha$ might have only brought about $\varphi$ conditionally, and other preceding actions that achieved those conditions must be considered as (indirect) cause of $\varphi$. Using this reasoning, in addition to the single action that brings about the effect of interest, one can also capture the chain of actions that build up to it. 
%\par
To formalize reasoning about effects, (KL21) introduced the notion of \emph{dynamic formulae}. %in the SC}. 
An effect $\varphi$ in their framework is thus a dynamic formula.\footnote{While (KL21) also study epistemic causation, we restrict our discussion to objective causality only.} 
% Also, as usual, we will often suppress the situation argument of $\varphi$. $\varphi[s]$ denotes the reintroduction of $s$ in $\varphi$; see below for a definition.} 
Given an effect $\varphi,$ the actual causes are defined relative to a {\em narrative} (variously known as a {\em scenario} or a {\em trace}) 
$s$. When $s$ is a ground situation, the tuple $\langle \mathcal{D},s,\varphi\rangle$ is often called a {\em causal setting} \cite{BatusovS18}. Also, it is assumed that $s$ is executable, and $\varphi$ was false before the execution of the actions in $s$, but became true afterwards, i.e.\ 
$\mathcal{D}\models \mathit{Executable}(s)\wedge\neg\varphi[S_0]\wedge\varphi[s]$. Here $\varphi$ is a formula with its situation arguments suppressed and $\varphi[s]$ denotes the formula obtained from $\varphi$ by restoring the given situation argument $s$ into all fluents in $\varphi$ (see Def.\ \ref{psiSAT} below).
%
%\emph{causal setting} that includes a theory $\mathcal{D}$ representing the domain dynamics, and a ground situation $\sigma$, representing the ``narrative'' (i.e.\ trace of events) where the effect was observed. 
%
%\begin{definition}[Causal Setting \cite{BatusovS18}]
%
%A causal setting is a tuple $\langle\mathcal{D},\sigma,$ $\Phi[s]\rangle$, where $\mathcal{D}$ is a theory, $\sigma$ is a ground 
%situation term of the form $do([\alpha_1,\cdots,\alpha_n], S_0)$ with ground action functions $\alpha_1,\cdots,\alpha_n$ such that $\mathcal{D}\models \mathit{executable}(\sigma)$, and $\Phi[s]$ 
%is a SC formula uniform in $s$ such that $\mathcal{D}\models\neg\Phi[S_0]\wedge\Phi[\sigma]$.
%
%\end{definition}
%As the theory $\mathcal{D}$ does not change, when referring to a causal setting we will often suppress $\mathcal{D}$ and simply write $\langle\sigma,\Phi\rangle$. 
%Also, here $\Phi$ is required to hold by the end of the narrative $\sigma,$ and thus we ignore the cases where $\Phi$ is not achieved by the actions in $\sigma$, since if this is the case, the achievement cause truly does not exist.   
\par
Note that since all changes in the SC result from actions, the potential causes of an effect $\varphi$ are identified with a set of action terms occurring in $s$. However, since $s$ might include multiple occurrences of the same action, one also needs to identify the situations where these actions were executed. To deal with this, (KL21) required that each situation be associated with a timestamp, which is an integer for their theory. 
Since in the context of knowledge, there can be different epistemic alternative situations (possible worlds) where an action occurs, using timestamps provides a common reference/rigid designator for the action occurrence. 
(KL21) assumed that the initial situation starts at timestamp 0 and each action increments the timestamp by one. Thus, their action theory includes the following axioms:
%
%\vspace{-4 mm}\begin{small}
%\begin{small}
\begin{eqnarray*}
%&&\mathit{Init}(s)\supset
&&\hspace{-.5 cm}\mathit{time}(S_0)=0,\\
&&\hspace{-.5 cm}\forall a,s,ts.\;\mathit{time}(do(a,s))=ts\equiv
%{}\\&&\hspace{9 mm}
\mathit{time}(s)=ts-1.
\end{eqnarray*}
%\end{small}
%\end{small}
%
%\vspace{-4 mm}
\noindent With this, causes in their framework is a non-empty set of action-timestamp pairs derived from the trace $s$ given $\varphi$. 
%
%*************************************************************************************************
%
\par
The notion of \emph{dynamic formulae} is defined as follows:
\begin{definition}
Let %$\Phi$ range over situation-suppressed formulae and 
$\vec{x}$, $\theta_a$, and $\vec{y}$ respectively range over object terms, action terms, and object and action variables. %, and $\theta_a$ range over action terms. 
The class of \emph{dynamic formulae} $\varphi$ is defined inductively using the following grammar:

\begin{small}
\begin{eqnarray*}
&& \hspace{-.7 cm}\varphi::=P(\vec{x}) \mid \mathit{Poss}(\theta_a)\mid\mathit{After}(\theta_a,\varphi)\mid\neg\varphi\mid\varphi_1\wedge\varphi_2{\color{black}\mid\exists\vec{y}.\varphi}
\end{eqnarray*}
\end{small}
\end{definition}

\noindent That is, a dynamic formula (DF) can be a situation-suppressed fluent, a formula that says that some action $\theta_a$ is possible, a formula that some DF holds after some action has occurred, or a formula that can built from other DF using the usual connectives. Note that $\varphi$ can have quantification over object and action variables, but we cannot have quantification over situations or mention the ordering over situations (i.e.\ $\sqsubset$). 
%Also, while it may include knowledge and intention modalities, $K$ or $G$-relations that do not come from the expansion of $\mathit{Know}$/$\mathit{Int}$ are not permitted. 
We will use $\varphi$ for DFs. 
\par
$\varphi[\cdot]$ is defined as follows:
\begin{definition}\label{psiSAT}
\begin{small}
\begin{eqnarray*}
&&  \hspace{-7 mm}\varphi[s]\defi
    \begin{cases}      
      P(\vec{x},s), & \!\!\textup{if }\varphi\textup{ is }P(\vec{x})\\
      \mathit{Poss}(\theta_a,s), & \!\!\textup{if }\varphi\textup{ is }\mathit{Poss}(\theta_a)\\
      \varphi'[do(\theta_a,s)], & \!\!\textup{if }\varphi\textup{ is }\mathit{After}(\theta_a,\varphi')\\
      \neg(\varphi'[s]), & \!\!\textup{if }\varphi\textup{ is }(\neg\varphi')\\
      \varphi_1[s]\wedge\varphi_2[s], & \!\!\textup{if }\varphi\textup{ is }(\varphi_1\wedge\varphi_2)\\
      \exists\vec{y}.\;(\varphi'[s]), & \!\!\textup{if }\varphi\textup{ is }(\exists\vec{y}.\;\varphi').\\
    \end{cases}      
\end{eqnarray*}
\end{small}
\end{definition}
%
\begin{comment}
\par
{\color{red}To facilitate reasoning about causes, we also define the following operator, which says that $\varphi$ was true before $\theta_a$.
\begin{definition}
    \[\mathit{Before}(\theta_a,\varphi)[s]\defi\exists s'.\;s=do(\theta_a,s')\wedge\varphi[s'].\]
\end{definition}
}
\end{comment}
%
\par
We will now present (KL21)'s definition of causes in the SC. The idea behind how causes are computed is as follows. Given an effect $\varphi$ and scenario $s$, if some action of the action sequence in $s$ triggers the formula $\varphi$ to change its truth value from false to true relative to $\mathcal{D}$, and if there are no actions in $s$ after it that change the value of $\varphi$ back to false, then this action is an actual cause of achieving $\varphi$ in $s$. Such causes are referred to as {\em primary} causes:%\footnote{Definitions \ref{PCause} and \ref{ACause} below are similar to those in \cite{KL21}, but adapted for HTSC by replacing the time-step of a situation with its starting time.}%444
\begin{definition}[Primary Cause (KL21)]\label{PCause} 
%\cite{KL21}]
\begin{small}
\begin{eqnarray*}
&&\hspace{-.7 cm}\mathit{CausesDirectly}(a,ts,\varphi,s)\defi\\
&& \exists s_a.\;\mathit{time}(s_a)=ts\wedge (S_0<do(a,s_a)\leq s)\wedge \\
&& \neg\varphi[s_a]\wedge\forall s'.(do(a,s_a)\leq s'\leq s\supset\varphi[s']).
\end{eqnarray*}
\end{small}
\end{definition} 
\noindent That is, $a$ executed at timestamp $ts$ is the \emph{primary cause} of effect $\varphi$ in situation $s$ 
iff $a$ was executed in a situation with timestamp $ts$ in scenario $s$, $a$ caused $\varphi$ to change its truth value to true, and no subsequent actions on the way to $s$ falsified $\varphi$. 
\par
%They showed that when used together with the single-step regression operator $\rho$, in addition to the single action that brings about the effect of interest, one can also capture the chain of actions that build up to it. 
Now, note that a (primary) cause $a$ might have been non-executable initially. Also, $a$ might have only brought about the effect conditionally and this context condition might have been false initially. Thus earlier actions in the trace that contributed to the preconditions and the context conditions of a cause must be considered as causes as well. The following definition captures both primary and indirect causes:\footnote{In this, we need to quantify over situation-suppressed DF. Thus we must encode such formulae as terms and formalize their relationship to the associated SC formulae. This is tedious but can be done essentially along the lines of \cite{GiacomoLL00}. We assume that we have such an encoding and use formulae as terms directly.} 
\begin{definition}[Actual Cause (KL21)]\label{ACause} %\cite{KL21}
\begin{small}
\begin{eqnarray*}
&&\hspace{-.7 cm}\mathit{Causes}(a,ts,\varphi,s)\defi{}\\
&&\hspace{-.7 cm}\forall P.[
\forall a,ts,s,\varphi.(\mathit{CausesDirectly}(a,ts,\varphi,s)\supset P(a,ts,\varphi,s))\\
&&\hspace{-.5 cm}{}\wedge\forall a,ts,s,\varphi.( \exists a'\!,ts'\!,s'\!.(\mathit{CausesDirectly}(a'\!,ts'\!,\varphi,s)\\
&&\hspace{19 mm}{}\land\mathit{time}(s')\!=\!ts'\land s'<s\\
&&\hspace{19 mm}{}\land P(a,ts,[\mathit{Poss}(a')\wedge\mathit{After}(a',\varphi)],s')\\
&&\hspace{13 mm}{}\supset P(a,ts,\varphi,s))\\
&&\hspace{-3 mm}{}]\supset P(a,ts,\varphi,s).
\end{eqnarray*}
\end{small}
\end{definition}
\noindent Thus, $\mathit{Causes}$ is defined to be the least relation $P$ such that if $a$ executed at timestamp $ts$ directly causes $\varphi$ in scenario $s$ then $(a,ts,\varphi,s)$ is in $P$, and if $a'$ executed at $ts'$ is a direct cause of $\varphi$ in $s$, the timestamp of $s'$ is $ts'$, $s'<s$, and $(a,ts,[\mathit{Poss}(a')\wedge\mathit{After}(a',\varphi)],s')$ is in $P$ (i.e.\ $a$ executed at $ts$ is a direct or indirect cause of $[\mathit{Poss}(a')\wedge\mathit{After}(a',\varphi)]$ in $s'$), then $(a,ts,\varphi,s)$ is in $P$. Here the effect $[\mathit{Poss}(a')\wedge\mathit{After}(a',\varphi)]$ is that $a'$ be executable and $\varphi$ hold after $a'$.
\par
The (KL21) formalization of actual causation was formulated for deterministic domains specified by BATs in the situation calculus.  However, it can be used directly for nondeterministic domains, for instance domains specified by NDBATs, as long as one focuses on scenarios that involve sequences of system actions, where both the agent actions and the environment reactions are known.  This is not surprising as NDBATs are special kinds of BATs and sequences of system actions are essentially situations. We illustrate this in the example below.
\paragraph{\textbf{\textup{Example (Cont'd).}}} 
Consider the system action sequence in the scenario $\sigma_1$, where
\begin{small}
\begin{eqnarray*}
&&\hspace{-8.5 mm}\sigma_1=do([\mathit{comm}(I_0, \mathit{Succ}),\mathit{move}(I_0, I_1, \mathit{NotVul}),\\
&&\hspace{5 mm}\mathit{move}(I_1, I_2, \mathit{Vul}),\mathit{move}(I_2, I_3, \mathit{NotVul})],S_0).
\end{eqnarray*}
\end{small}
%\end{eqnarray*}
$\!$We are interested in computing the causes of the effect $\varphi_1=\mathit{Vul}$, i.e., the robot becoming vulnerable, in this scenario $\sigma_1$.
%\textbf{[YL Drop or fix!:
%Then, according to Definition $\ref{PCause}$, the causal setting $\langle \mathcal{D}_1,\sigma_1, \psi_1 \rangle$ satisfies the achievement condition $\psi_1$ via the following situation term:\\
%$\text{comm}(I_0, \mathit{Success});\text{move}(I_0, I_1, \mathit{NotVul});\text{move}(I_1, I_2, \mathit{Vul});\text{move}(I_2, I_3, \mathit{e})$.}\\
It can be shown that:
\begin{proposition}[Causes of $\varphi_1$ in $\sigma_1$]\label{EProp1}
\begin{small}
\begin{eqnarray*}
&&\hspace{-7 mm}\mathcal{D}_1\models\neg\mathit{Causes}(\mathit{comm}(I_0,\mathit{Succ}),0,\varphi_1,\sigma_1)\\
&&{}\land\mathit{Causes}(\mathit{move}(I_0,I_1,\mathit{NotVul}),1,\varphi_1,\sigma_1)\\
&&{}\land\mathit{Causes}(\mathit{move}(I_1,I_2,\mathit{Vul}),2,\varphi_1,\sigma_1)\\
&&{}\land\neg\mathit{Causes}(\mathit{move}(I_2,I_3,\mathit{NotVul}),3,\varphi_1,\sigma_1).
\end{eqnarray*}
\end{small}
\end{proposition}
Thus, e.g., the action $\mathit{move}(I_1,I_2,\mathit{Vul})$ executed at timestamp $2$ is a cause since it directly caused the robot to become vulnerable. 
Moreover, $\mathit{move}(I_0,I_1,\mathit{NotVul})$ executed at timestamp $1$ can be shown to be an indirect cause of the $\varphi_1$. This is because by axioms ($1$) and ($1'$) the primary cause of moving from location $I_1$ to $I_2$ i.e.\ $\mathit{move}(I_1,I_2,\mathit{Vul})$ is only possible when the robot is at $I_1$, which in this scenario was brought about by $\mathit{move}(I_0,I_1,\mathit{NotVul})$.
\section{Agent Actions as Causes in the NDSC}%Nondeterministic Situation Calculus}
We now turn our attention to causation in nondeterministic domains.
As mentioned, when the scenario is a sequence of system actions where both the agent actions and the environment reactions are specified, we can use the (KL21) formalization presented earlier to reason about actual causation, and identify causes for effects that are system actions containing both agent actions and environment reactions.
But in many cases, we would like to consider scenarios that are sequences of agent actions only and where we don't know what the environment reactions were.  Moreover, we want to analyse which agent actions were causes of given effects independently of the environment reactions.
%
% In such domains, it is possible that the agent is only aware of the sequence of agent actions that occurred before the effect was observed, i.e.\ the actual situation resulted from this execution (that also includes choices made by the environment) may or may not be known.
%
We address this question in this section.

We start by defining
%As such, we need to define 
a notion of \emph{nondeterministic causal setting} that generalizes causal settings and reflects the agent's ignorance about the environment's choices.
\begin{definition}[Nondeterministic Causal Setting]\label{NDSetting}
A nondeterministic cau\-sal setting is a tuple $\langle\mathcal{D},\vec{\alpha},\varphi\rangle$, where $\mathcal{D}$ is a NDBAT, $\vec{\alpha}$ is a sequence of agent actions representing the nondeterministic scenario, and $\varphi$ is a dynamic formula s.t.: 
\[\mathcal{D}\models\neg\varphi[S_0]\wedge\mathit{PAfter}(\vec{\alpha},\varphi,S_0).\]
\end{definition}
Thus a scenario in a nondeterministic causal setting (ND setting, henceforth) is modeled using a sequence of agent actions $\vec{\alpha}$, with the assumption that this sequence was executed starting in $S_0$. {\color{CBRed}Also, $\varphi$ was initially false and possibly holds after $\vec{\alpha}$ starting in $S_0$, i.e.\ holds after at least one execution of $\vec\alpha$ starting in $S_0$. Notice that, since we deal with actual causation, which is the problem of determining the causes of an {\em already observed} effect, just like most previous work on actual causation, we also assume that the effect has already been observed after the (in our case, nondeterministic) actions in the scenario occurred. Thus, any execution after which the effect $\varphi$ did not occur is excluded from consideration and the agent only considers executions after which $\varphi$ holds to be candidates for the actual one (as the agent has already observed the effect).}
\par
As before, in our framework causes are action and timestamp pairs. However, these actions are now agent actions. %\footnote{Note that since system actions in NDSC are deterministic in nature, causation relative to a scenario built from a sequence of these actions can be defined exactly as in Definition \ref{Causes}.} 
Also, since each of the agent actions in the scenario can have multiple outcomes, depending on these outcomes, we might sometimes call an agent action a cause and sometimes not a cause. 
In some cases, an agent action is a cause of an effect for all possible environment choices associated with the actions in the scenario. 
%
%Put otherwise, 
In general, given a scenario which is a sequence of agent actions, we will get a tree of possible executions, where each branch is the execution produced by a given set of environment reactions to the agent actions in the sequence.
% the execution of agent actions in the scenario starting in the initial situation $S_0$ yields a situation tree grounded in $S_0$.
In this tree, it might be that only on certain branches a system action associated with an agent action executed at some timestamp is a cause. Additionally, it is also possible that all the system actions associated with this agent action is a cause in their respective branch. Thus, we have to define two notions of actual causes for agent actions in nondeterministic domains, namely \emph{possibly causes} and \emph{certainly causes}:\footnote{We allow both agent actions and system actions to be viewed as causes. An additional possibility is to view the environment's choices as causes, for instance, when there are no other ways of achieving an effect but via certain environment reactions. However, the consequences of such a definition is not clear. For instance, is it reasonable to assign responsibility/blame to nature? Rather than engaging in such philosophical questions, in this paper we focus on causes that involve the agent.}
\begin{definition}[Possibly Causes]\label{PossiblyCauses}
Let $\langle\mathcal{D},\vec{\alpha},\varphi\rangle$ be a ND setting and $\beta(\vec{x})$ an agent action in $\vec{\alpha}$.
\begin{small}
\begin{eqnarray*}
&&\hspace{-.7 cm}\mathit{PCauses}(\beta(\vec{x}),t,\varphi,\vec{\alpha})\defi\mbox{}\\
&&\hspace{-.2 cm}\exists s.\;\mathit{Do}_\mathit{ag}(\vec{\alpha}, S_0, s)\wedge\varphi[s]\wedge\exists e.\;\mathit{Causes}(\beta(\vec{x},e),t,\varphi,s).
\end{eqnarray*}
\end{small}
\end{definition}
That is, agent action $\beta(\vec{x})$ executed at timestamp $t$ possibly causes $\varphi$ in scenario $\vec\alpha$ iff there is an execution of $\vec\alpha$ that reaches some situation $s$, $\varphi$ holds in $s$, and for some environment reaction $e$ the associated system action $\beta(\vec{x},e)$ executed at timestamp $t$ is a (deterministic) cause of $\varphi$ in scenario $s$.
\begin{definition}[Certainly Causes]\label{certainlyCauses}
Let $\langle\mathcal{D},\vec{\alpha},\varphi\rangle$ be a ND setting and $\beta(\vec{x})$ an agent action in $\vec{\alpha}$.
\begin{small}
\begin{eqnarray*}
&&\hspace{-.7cm}\mathit{CCauses}(\beta(\vec{x}),t,\varphi,\vec{\alpha})\defi\mbox{}\\
&&\hspace{-.2cm}\forall s.\;\mathit{Do}_\mathit{ag}(\vec{\alpha},S_0,s)\wedge\varphi[s]\supset\exists e.\;\mathit{Causes}(\beta(\vec{x},e),t,\varphi,s).
\end{eqnarray*}
\end{small}
\end{definition}
Thus agent action $\beta(\vec{x})$ executed at timestamp $t$ certainly causes $\varphi$ in scenario $\vec\alpha$ iff for all executions of $\vec\alpha$ that reach some situation $s$ in which $\varphi$ holds, there is a environment reaction $e$ such that the associated system action $\beta(\vec{x},e)$ executed at timestamp $t$ is a (deterministic) cause of $\varphi$ in scenario $s$. {\color{CBRed}Note that this does not require a system action associated with $\beta(\vec{x})$ to be a cause in executions $s$ where $\varphi$ do not hold; this is because since the agent is assumed to have observed the effect $\varphi$, such executions can be ruled out as unrealistic, i.e.\ inconsistent with this assumption.}
\paragraph{\textbf{\textup{Example (Cont'd).}}} 
\begin{figure}[t]
\begin{center}
\resizebox{0.3\textwidth}{!}{ % Adjust the 0.8\textwidth to your desired width
\begin{tikzpicture}
\node (v1) at (0,3.5) {\(S_0, \neg v\)};
\node (v2) at (0,2.5) {\(S_1, \neg v\)};
\draw  (v1) edge (v2);
\node (v3) at (-2,1.5) {\(S_2^a, v\)};
\node (v4) at (2,1.5) {\(S_2^b, \neg v\)};
\draw  (v2) edge (v3);
\draw  (v2) edge (v4);
\node (v6) at (-0.5,0.5) {\(S_{31}^b, v\)};
\node (v5) at (-3.5,0.5) {\(S_{31}^a, v\)};
\node (v7) at (0.5,0.5) {\(S_{32}^a,  v\)};
\node (v8) at (3.5,0.5) {\(S_{32}^b, \neg v\)};
\draw  (v3) edge (v5);
\draw  (v3) edge (v6);
\draw  (v4) edge (v7);
\draw  (v4) edge (v8);
\node (v9) at (-3.5,-0.5) {\(S_{41}^b, v\)};
\node (v10) at (-0.5,-0.5) {\(S_{42}^b, v\)};
\node (v11) at (0.5,-0.5) {\(S_{43}^b, v\)};
\node (v12) at (3.5,-0.5) {\(S_{44}^b, \neg v\)};
\draw  (v5) edge (v9);
\draw  (v6) edge (v10);
\draw  (v7) edge (v11);
\draw  (v8) edge (v12);
\end{tikzpicture}
}
\end{center}
\mbox{}\vspace{-5 mm}~
\caption{Executions of agent action sequence $\protect\vec{\alpha_1}$.}\label{fig:treeExecs}
\end{figure}
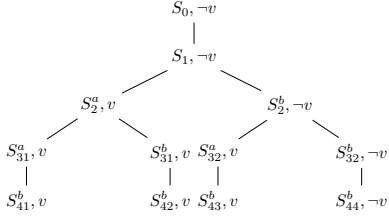
Consider the agent action sequence 
\begin{small}
\begin{eqnarray*}
\vec{\alpha_1}=\mathit{comm}(I_0);\mathit{move}(I_0,I_1);\mathit{move}(I_1,I_2);\mathit{move}(I_2, I_3).
\end{eqnarray*}
\end{small}
$\!\!\!\!$When executed starting in $S_0$, $\vec{\alpha_1}$ produces the tree of possible executions shown in Fig.~\ref{fig:treeExecs}. Here, the superscripts $a$ and $b$ represent environment choices $\mathit{Vul}$ and $\mathit{NotVul}$, respectively, and $v$/$\neg v$ indicates whether the agent has become vulnerable or not. Thus, e.g., in this tree $S_1=do(\mathit{comm}(I_0,\mathit{Succ}),S_0)$ and $S_2^a=do(\mathit{move}(I_0,I_1,\mathit{Vul}),S_1)$, etc. 
It is easy to see that %the ND causal setting $\langle\mathcal{D}_1,\vec{\alpha_1}, \varphi_1\rangle$ 
the execution of $\vec{\alpha_1}$ starting in $S_0$ possibly satisfies $\varphi_1 = \mathit{Vul}()$, e.g. due to the existence of path $(S_0,S_{43}^b)$ in this tree over which the system action sequence in $\sigma_1$ is executed. Given ND causal setting $\langle \mathcal{D}_1,\vec{\alpha}_1,\varphi_1\rangle$, we can in fact show that: 
%
%The ND causal setting $\langle \mathcal{D}_1,\vec{\alpha}_1, \psi_1\rangle$ possibly satisfies $\psi_1 = Vul(s)$, i.e., the robot becomes vulnerable, starting from $S_0$.
%
\begin{proposition}
% Assume that $\mathcal{D}_1$ be a set of axioms for the robot navigating between different locations. Then the following lists the causes of $\psi_1$ in $\Vec{\alpha}_1$.
\begin{small}
\[\begin{array}{l}
\hspace{2 mm}\mathcal{D}_1\models\neg\mathit{PCauses}(\mathit{comm}(I_0),0,\varphi_1,\vec{\alpha_1})\\
\hspace{2 em}{}\wedge\mathit{CCauses}(\mathit{move}(I_0, I_1),1,\varphi_1,\vec{\alpha_1})\\
%
%\hspace{2 em}{}\wedge\mathit{PCauses}(\mathit{move}(I_0,I_1),1,\varphi_1,\vec{\alpha_1})\\
%
\hspace{2 em}{}\wedge\mathit{PCauses}(\mathit{move}(I_1,I_2),2,\varphi_1,\vec{\alpha_1})\\
\hspace{2 em}{}\wedge\neg\mathit{CCauses}(\mathit{move}(I_1,I_2),2,\varphi_1,\vec{\alpha_1})\\
\hspace{2 em}{}\wedge\neg\mathit{PCauses}(\mathit{move}(I_2,I_3),3,\varphi_1,\vec{\alpha_1}).
\end{array}\]
\end{small}
\end{proposition}
Thus, $\mathit{comm}(I_0)$ and $\mathit{move}(I_2,I_3)$ do not possibly causes $\varphi_1$ because they are not a cause (in the deterministic sense) of $\varphi_1$ in any of the executions of $\vec{\alpha_1}$ depicted in Fig.\ \ref{fig:treeExecs} in which $\varphi_1$ holds, i.e.\ in scenarios $S_{41}^b,S_{42}^b,$ and $S_{43}^b$. 
Moreover, $\mathit{move}(I_0,I_1)$ certainly causes (and thus also possibly causes) $\varphi_1$; this is because in all the executions of $\vec{\alpha_1}$ starting in $S_0$ over which $\varphi_1$ holds, it is either a direct cause of $\varphi_1$ (as in situations $S^b_{41}$ and $S^b_{42}$), or it is an indirect cause of $\varphi_1$ (as in $S^b_{43}$). To see the latter, recall Proposition \ref{EProp1} above.  
%
%within $\vec{\alpha}_1$. It is possibly a cause of $\psi_1$ because in some executions of $\vec{\alpha}_1$ starting from $S_0$, the robot becomes vulnerable as a result of the system action $\mathit{move}(I_0, I_1, \mathit{Vul})$, e.g., in $s_2^a$, but not in $s_2^b$.
%%where $\mathit{move}(I_0, I_1, \mathit{Vul})$ is a cause of vulnerability. 
%But it is certainly a cause of $\psi_1$ because in all execution of $\vec{\alpha}_1$ starting from $S_0$, either the robot gets vulnerable as in $s_2^a$ because system action $\mathit{move}(I_0, I_1, \mathit{Vul})$ occurs, or , as in $s_2^b$, it is a secondary cause of the system action $\mathit{move}(I_1, I_2, \mathit{Vul})$ which make the robot vulnerable as it makes its precondition of having the robot in location $I_1$ satisfied.
%
Finally, $\mathit{move}(I_1, I_2)$ possibly causes $\varphi_1$, but not certainly causes it; this is because there is at least one execution of $\vec{\alpha_1}$ in which it is a cause of $\varphi_1$, e.g.\ the one ending in $S_{43}^b$, but 
it is not in
all executions in which $\varphi_1$ holds, e.g.\ in %scenario 
$S_{41}^b$.

\section{Reasoning about Achievement Causes}
%{\color{red}
We now extend regression to answer queries about various notions of causes, and $\mathit{CAfter}$ and $\mathit{PAfter}$. Note that, combining ideas from (KL21) and (DL21), recently in a preliminary work \cite{RKLY24} showed that one can obtain equivalent regressable formulae for Possibly Causes and Certainly Causes. 
%{\color{blue} 
However, 
they do this by translating these into formulae that refer to all possible executions of the scenario/agent action sequence, 
%to do this, they employed a translation-based approach that looks at all possible evolutions of the scenario, 
making reasoning intractable. 
Here, we develop a more effective approach by extending regression so that Certainly/Possibly Causes in $do(a,s)$ is transformed into an equivalent formula about $s$.
%by extending regression propose a method that is more convenient compared to previous approaches and does not require examining every possible execution.
%[Yves: say a bit more (along the lines of the introduction).
%]
%}
%regression of $\mathit{Causes}$, $\mathit{PAfter/CAfter}$, and $\mathit{PCauses/CCauses}$.
%}
In the following, we prove a series of properties, which will be the basis of our proposed extended regression operator.
\paragraph{\textup{\textbf{Reasoning about System Actions as Causes}}.} We start by showing that $\mathit{Causes}$ in $do(a,s)$ can be reduced to a formula that only mentions $\mathit{Causes}$ in $s$:
%
\begin{comment}
\begin{theorem}\label{Causes to FO-F}
If $do(a, s)$ is a ground situation term, $b$ is a ground system action
term, and $t$ is an integer, then
\begin{small}
\begin{equation*}
\begin{array}{l}
\mathcal{D}\models\mathit{CausesDirectly}(b,t,\varphi,do(a,s))\\
{}\equiv(\mathit{time}(s) = t \land b = a \land\lnot \varphi[s] \land \varphi(do(a, s))\vee{}\\
(\mathit{time}(s) > t \land \varphi[s] \land \varphi[do(a, s)] \land 
\mathit{CausesDirectly}(b,t,\varphi,s))\\
{}\equiv(\mathit{time}(s) = t \land b = a \land\lnot \varphi[s] \land \mathcal{R}[(\varphi, a)])\vee{}\\
(\mathit{time}(s) > t \land \varphi[s] \land \mathcal{R}[(\varphi, a)] \land 
\mathit{CausesDirectly}(b,t,\varphi,s)).
\end{array}
\end{equation*}
\end{small}
{\color{red}[fix the num of arguments of $\mathcal{R}$, 1 vs. 2].}
\end{theorem}
\end{comment}
%First equivalency says that We have two cases either $time(s)=t$ and $b$ is the last action in the action sequence $(b = a)$ and before performing $b$ the formula $\varphi$ does not hold and it becomes true immediately after performing $b$ or $time(s)>t$ and before performing the last action $a$ in action sequence the formula $\varphi$ became true, in fact $b$ is a directly cause of $\varphi$ in time $t< n-1$.\\
%Now we demonstrate that the definition of $\mathit{Causes}$ can be expanded to the set of first order formulas and then it is possible to apply the regression operator to these formulas as follows.
\begin{proposition}
\label{CausesProp}
% Let $\mathcal{D}$ be an  NDBAT, $do(a, s)$ be a ground situation term, $b$ be a ground system action
% term, $\varphi$ be a ground dynamic formula (i.e.\ it does not include action variables), and $t$ be an integer {\color{blue} [need ground?]}. Then we have:
%{\color{red}
\begin{small}
\begin{equation*}
\begin{array}{l}
\hspace{-.3 cm}\mathcal{D}\models\mathit{Causes}(b,t,\varphi,do(a,s))
\equiv{}\\
\hspace{-.2 cm}(\mathit{time}(s) = t \land b = a \land\lnot \varphi[s] \land \varphi[do(a,s)])\vee{}\\
\hspace{-.2 cm}(\mathit{time}(s) > t \land \varphi[s] \land \varphi[do(a,s)] \land 
\mathit{Causes}(b,t,\varphi,s))\vee{}\\
\hspace{-.2 cm}(\mathit{time}(s) > t\wedge\lnot \varphi[s] \land \varphi[do(a,s)]\\
\hspace{1.8 cm} {}\land\mathit{Causes}(b,t,\mathit{Poss}(a) \land\mathit{After}(a,\varphi),s)).%\equiv{}\\
%
%\hspace{-.2 cm}(\mathit{time}(s) = t \land b = a \land\lnot \varphi[s] \land \mathcal{R}[\varphi[do(a,s)]])\vee{}\\
%\hspace{-.2 cm}(\mathit{time}(s) > t \land \varphi[s] \land \mathcal{R}[\varphi[do(a,s)]] \land 
%\mathit{Causes}(b,t,\varphi,s))\vee{}\\
%\hspace{-.2 cm}(\mathit{time}(s) > t\wedge\lnot \varphi[s] \land \mathcal{R}[\varphi[do(a,s)]]\\
%\hspace{1.8 cm}{}\land\mathit{Causes}(b,t,\mathit{Poss}(a) \land\mathcal{R}[\varphi[do(a,s')]]^{-1},s)).
\end{array}
\end{equation*}
\end{small}
%%}
\end{proposition}

\noindent
{\color{CBRed}\textbf{Proof Sketch:}
$\Leftarrow$ For the first disjunct, the result follows immediately from the first implication in the definition of $Causes$.  For the third disjunct, it follows immediately from the second implication in the definition of $Causes$.  For the second disjunct, the result follows by induction on the number of actions in the causal chain from $b$ to the direct cause of $\varphi$ holding in $do(a,s)$.  

$\Rightarrow$ The result follows by induction on the number of actions in the causal chain from $b$ to the direct cause of $\varphi$ holding in $do(a,s)$.
$\Box$}

\noindent Thus, $\mathit{Causes}(b,t,\varphi,do(a,s))$ can be reduced to one of 3 cases. Either $b$, which is the same action as $a$, directly causes $\varphi$ in $do(a,s)$, and in that case the timestamp of $s$ is $t$ (i.e.\ $a$ was performed at $t$), $\varphi$ was false before the execution of $a$, and became true afterwards; or $b$ caused $\varphi$ in an earlier situation and $a$ has no (positive or negative) contributions to this, and in that case the timestamp of $s$ (i.e.\ the execution time of $a$) is greater than $t$, $\varphi$ was true before and after the execution of $a$, and $b$ executed at $t$ caused $\varphi$ in the preceding situation $s$; or $a$ directly caused $\varphi$ and $b$ indirectly caused $\varphi$ by ensuring that the preconditions of $a$ and the context conditions under which $a$ caused $\varphi$ holds, and in that case the timestamp of $s$ is greater than $t$, $\varphi$ was false before the execution of $a$ and became true afterwards, and $b$ executed at $t$ caused the effect $[\mathit{Poss}(a)\land\mathit{After}(a,\varphi)]$ in scenario $s$. 
\paragraph{\textup{\textbf{Reasoning about Certainly and Possibly Causes.}}} Next, we show how a $\mathit{CAfter}$ formula (and a $\mathit{PAfter}$ formula) relative to a sequence of agent actions starting in some situation can be reduced to one relative to the shorter sequence obtained by removing the last action from the sequence. This result along with the next will be then used for defining regression for Certainly/Possibly Causes.
%
%Now We aim to show that $\mathit{PCauses}$ and $\mathit{CCauses}$ can be converted to first order formulas and then we can apply regression operation to these first order formulas. To achieve this, we need to have the following theorems and proposition. By definition of $\mathit{PAfter}$ and $\mathit{CAfter}$, it is possible to have:
\begin{proposition}\label{CAfterProp}
% Let $\mathcal{D}$ be an NDBAT, $\alpha_1,\cdots,\alpha_n$ be ground agent actions, $\varphi$ be a ground dynamic formula, and $s$ be a ground situation {\color{blue} [need ground?]}. Then we have:
%
%{\color{blue}
\begin{small}
\begin{equation*}
\begin{array}{l}
\hspace{-.5 cm}\mathcal{D} \models\mathit{CAfter}([\alpha_1,\ldots,\alpha_{n-1},\alpha_n], \varphi,s) \equiv{}\\
\hspace{-.2 cm}\mathit{CAfter}([\alpha_1,\ldots,\alpha_{n-1}],
\mathit{CAfter}(\alpha_n, \varphi),s)\equiv{}\\
\hspace{-.2 cm}\mathit{CAfter}([\alpha_1,\ldots,\alpha_{n-1}],\\
\hspace{0.5 cm}\forall e.\;\mathit{Poss}(\alpha_n(e),\mathit{now})\supset \varphi[do(\alpha_n(e),\mathit{now})],s).\\
%\mathcal{R}[\varphi[do(\alpha_n(e),s')]]^{-1}, s),\\
\\
\hspace{-.5 cm}\mathcal{D} \models\mathit{PAfter}([\alpha_1,\ldots,\alpha_{n-1},\alpha_n], \varphi,s) \equiv{}\\
\hspace{-.2 cm}\mathit{PAfter}([\alpha_1,\ldots,\alpha_{n-1}], \mathit{PAfter}(\alpha_n, \varphi),s) \equiv{}\\
\hspace{-.2 cm}\mathit{PAfter}([\alpha_1,\ldots,\alpha_{n-1}], \\
\hspace{0.5 cm}\exists e.\;\mathit{Poss}(\alpha_n(e),\mathit{now})\wedge \varphi[do(\alpha_n(e),\mathit{now})],s).
%\mathcal{R}[\varphi[do(\alpha_n(e),s')]]^{-1}, s).
\end{array}
\end{equation*}
\end{small}
%%}
%
%{\color{blue}[Perhaps we should get rid of $\mathcal{R}$ here; later we use $\mathcal{R}_{ext}$]}
\end{proposition}
Notice that the situation argument in the formula inside the context of $\mathit{CAfter}$ (and $\mathit{PAfter}$) above will be provided by the external context once the $\mathit{CAfter}$ (and $\mathit{PAfter}$, resp.) is fully expanded. We often suppress this situation argument; we also sometimes use a placeholder $\mathit{now}$ that stands for it. 
\par
Finally, we can show that a $\mathit{CCauses}$ (and $\mathit{PCauses}$) formula can be converted to a certainly after (possibly after, resp.) causes formula. 
%Also by the definition of $\mathit{PCauses}$ and $\mathit{CCauses}$, we have:
\begin{proposition}\label{CCausesProp}
% Let $\mathcal{D}$ be an NDBAT, $\beta$ be a ground agent action term, $\vec\alpha$ be a ground agent action sequence, $\varphi$ be a ground dynamic formula, and $t$ be an integer {\color{blue} [need ground?]}. Then we have:
%{\color{red}
\begin{small}
\begin{equation*}
\begin{array}{l}
\hspace{-7 mm}\mathcal{D}\models\mathit{CCauses}(\beta,t,\varphi,\vec{\alpha})\equiv\\
\hspace{0 mm}\mathit{CAfter}(\vec{\alpha},\neg\varphi\vee\exists e.\;\mathit{Causes}(\beta(e),t,\varphi),S_0).\\
\\
\hspace{-7 mm}\mathcal{D}\models\mathit{PCauses}(\beta,t,\varphi,\vec{\alpha})\equiv\\
\hspace{0 mm}\mathit{PAfter}(\vec{\alpha},
\varphi\wedge\exists e.\;\mathit{Causes}(\beta(e),t,\varphi),S_0).
\end{array}
\end{equation*}
\end{small}
%%}
\end{proposition}
\paragraph{The Extended Regression Operator.}
We extend the set of regressable formulae to include $\mathit{Causes}(b,t,\varphi,do(a,s))$, $\mathit{CAfter}(\vec\alpha,\varphi,s)$, $\mathit{PAfter}(\vec\alpha,\varphi,s)$, $\mathit{CCauses}(\beta,t,\varphi,\vec{\alpha})$, and $\mathit{PCauses}(\beta,t,\varphi,\vec{\alpha})$, {\color{black} with the same restrictions on their arguments as imposed in
the original definition of regressable formula,}
%Proposition \ref{CausesProp} to \ref{CCausesProp}, 
as well as $\mathit{CAfter}(\epsilon,\varphi,s)$ and $\mathit{PAfter}(\epsilon,\varphi,s)$, where $\epsilon$ is the empty agent action sequence.
\par
We also extend $\mathcal{R}$ to define regression of these additional constructs; regression for the other cases are covered by Definition \ref{regressionDef} above. We denote this one-step extended regression operator using $\mathcal{R}_{ext}$. As usual, we use $\mathcal{R}^*_{ext}$ to denote the repeated application of $\mathcal{R}_{ext}$ until further applications leave the argument formula unchanged. %The additional cases of $\mathcal{R}_{ext}$ are as follows.
\begin{definition}[The Extended Regression Operator $\mathcal{R}_{ext}$]
\mbox{}
\begin{itemize}
\item[](1) -- (3) When the formula $\phi$ to be regressed is of the forms discussed in Definition \ref{regressionDef}, $\mathcal{R}_{ext}(\phi)=\mathcal{R}(\phi)$.
\item[](4) If $\phi$ is an extended regressable formula of the form $\mathit{Causes}(b,t,\varphi,do(a,s))$, then: 
%{\color{red}
\begin{small}
\begin{eqnarray*}
&&\hspace{-7 mm}\mathcal{R}_{ext}[\mathit{Causes}(b,t,\varphi,do(a,s))]={}\\
%&&(\mathit{time}(s) = t \land b = a \land\lnot \varphi[s] \land \mathcal{R}[(\varphi, a)])\vee\\
%&&(\mathit{time}(s) > t \land \varphi[s] \land \mathcal{R}[(\varphi, a)]\land\\
%&&\hspace{4 cm} \mathit{Causes}(b,t,\varphi,s))\vee \\
%&&(\mathit{time}(s) > t\wedge\lnot \varphi[s] \land \mathcal{R}[(\varphi, a)] \land\\
%&&\hspace{1.5 cm} \mathit{Causes}(b,t,\mathit{Poss}(a) \land\mathit{After}(a,\varphi),s)).\\
&&\hspace{-7 mm}(\mathit{time}(s) = t \land b = a \land\lnot \varphi[s] \land\mathcal{R}_{ext}[\varphi[do(a,s)]])\vee{}\\
&&\hspace{-.7 cm}(\mathit{time}(s) > t \land \varphi[s] \land \mathcal{R}_{ext}[\varphi[do(a,s)]] \land 
\mathit{Causes}(b,t,\varphi,s))\vee{}\\
&&\hspace{-.7 cm}(\mathit{time}(s) > t\wedge\lnot \varphi[s] \land \mathcal{R}_{ext}[\varphi[do(a,s)]]\\
&&\hspace{0.5 cm}{}\land\mathit{Causes}(b,t,\mathit{Poss}(a) \land\mathcal{R}_{ext}[\varphi[do(a,s')]]^{-1},s)).
\end{eqnarray*}
\end{small}
%%}
%
\item[](5)
If $\phi$ is an extended regressable formula that is of the form $\mathit{CAfter}(\vec\alpha, \varphi,s)$ or $\mathit{PAfter}(\vec\alpha, \varphi,s)$ with a possibly empty sequence $\epsilon$ of agent actions $\vec\alpha$, then:
\begin{small}
%\begin{eqnarray*}
%&& \mathcal{R}[\varphi]\equiv\mathit{CAfter}([\alpha_1, \dots, \alpha_{n-1}], \\
%&& \forall e.\mathit{Poss}(\alpha_n(\vec{x}, e))\supset \mathcal{R}[(\varphi, A_n(\vec{x}, e'))], s).\\
%&& \mathcal{R}[\varphi]\equiv\mathit{PAfter}([\alpha_1, \dots, \alpha_{n-1}], \\
%&& \exists e.\mathit{Poss}(\alpha_n(\vec{x}, e))\wedge \mathcal{R}[(\varphi, A_n(\vec{x}, e'))], s).
%\end{eqnarray*}
%{\color{red}
\begin{eqnarray*}
&&\hspace{-.5 cm}\mathcal{R}_{ext}[\mathit{CAfter}(\epsilon,\varphi,s)] = \mathcal{R}_{ext}[\varphi[s]].\\
%&&\\
&&\hspace{-.5 cm}\mathcal{R}_{ext}[\mathit{CAfter}([\alpha_1,\ldots,\alpha_{n-1},\alpha_n],\varphi,s)] ={}\\
&&\hspace{-.2 cm}\mathit{CAfter}([\alpha_1,\ldots,\alpha_{n-1}],\\
&&\hspace{0.5 cm}\forall e.\;\mathit{Poss}(\alpha_n(e))\supset \mathcal{R}_{ext}[\varphi[do(\alpha_n(e),s')]]^{-1}, s).\\
%&&\\
&&\hspace{-.5 cm}\mathcal{R}_{ext}[\mathit{PAfter}(\epsilon,\varphi,s)] = \mathcal{R}_{ext}[\varphi[s]].\\
%&&\\
&&\hspace{-.5 cm}\mathcal{R}_{ext}[\mathit{PAfter}([\alpha_1,\ldots,\alpha_{n-1},\alpha_n],\varphi,s)] ={}\\
&&\hspace{-.2 cm}\mathit{PAfter}([\alpha_1,\ldots,\alpha_{n-1}],\\
&&\hspace{0.5 cm}\exists e.\;\mathit{Poss}(\alpha_n(e))\land\mathcal{R}_{ext}[\varphi[do(\alpha_n(e),s')]]^{-1}, s).\\
\end{eqnarray*}
%%}
\end{small}
\item[](6) If $\phi$ is an extended regressable formula that is of the form $\mathit{CCauses}(\beta,t,\varphi,\vec{\alpha})$ or $\mathit{PCauses}(\beta,t,\varphi,\vec{\alpha})$, then:
\begin{small}
%{\color{red}
\begin{eqnarray*}
%&&\hspace{-.7cm}\mathcal{R}[\mathit{CCauses}(\beta(\vec{x}),t,\varphi,\alpha_1,\cdots,\alpha_{n-1},\alpha_n)]\equiv \\
%&& \hspace{-.3 cm}\mathit{CAfter}([\alpha_1,\ldots,\alpha_{n-1}], \forall e'. \mathit{Poss}(A_{n}(\vec{x}, e'), now) \supset,\\
%&& \lnot \mathcal{R}[\varphi(\varphi, A_n(\vec{x}, e'))] \lor \\
%&& (\varphi[now] \land \exists e. Causes(\beta(\vec{x},e),t,\varphi)[now])\lor {}\\
%&& (\mathcal{R}[\varphi(\varphi, A_n(\vec{x}, e'))]\wedge \neg \varphi \land \exists e. Causes(\beta(\vec{x},e).t, \\
%&& [\mathit{Poss}(A_n(\vec{x}, e'))\wedge \mathcal{R}[(\varphi, A_n(\vec{x}, e'))]])), S_0).
%
&&\hspace{-.7cm}\mathcal{R}_{ext}[\mathit{CCauses}(\beta,t,\varphi,[\alpha_1,\cdots,\alpha_{n-1},\alpha_n])]=\\
&& \hspace{-.3 cm}\mathit{CAfter}([\alpha_1,\ldots,\alpha_{n-1}],\\
&&\forall e'.\;\mathit{Poss}(\alpha_{n}(e'))\supset\mathcal{R}_{ext}[\phi^*[do(\alpha_n(e'),s')]]^{-1},S_0),\\
%&&\\
&&\hspace{-.7cm}\mathrm{where,}\;\;\phi^*=\neg\varphi\vee\exists e.\;\mathit{Causes}(\beta(e),t,\varphi).
\end{eqnarray*}
%%}
%{\color{red} [Add missing situation??]}
\end{small}
%
%{\color{red}
\begin{small}
\begin{eqnarray*}
%&&\hspace{-.7cm}\mathcal{R}[\varphi]\equiv \\
%&& \hspace{-.3 cm}\mathit{PCauses}([\alpha_1,\ldots,\alpha_{n-1}], \exists e'. \mathit{Poss}(A_{n}(\vec{x}, e'), now) \wedge,\\
%&& \lnot \mathcal{R}[\varphi(\varphi, A_n(\vec{x}, e'))] \lor \\
%&& (\varphi[now] \land \exists e. Causes(\beta(\vec{x},e),t,\varphi)[now])\lor {}\\
%&& (\mathcal{R}[(\varphi, A_n(\vec{x}, e'))]\wedge \neg \varphi \land \exists e. Causes(\beta(\vec{x},e).t, \\
%&& [\mathit{Poss}(A_n(\vec{x}, e'))\wedge \mathcal{R}[(\varphi, A_n(\vec{x}, e'))]])), S_0).
%
&&\hspace{-.7cm}\mathcal{R}_{ext}[\mathit{PCauses}(\beta,t,\varphi,[\alpha_1,\cdots,\alpha_{n-1},\alpha_n])]=\\
&& \hspace{-.3 cm}\mathit{PAfter}([\alpha_1,\ldots,\alpha_{n-1}],\\
&&\exists e'.\;\mathit{Poss}(\alpha_{n}(e'))\land\mathcal{R}_{ext}[\phi^*[do(\alpha_n(e'),s')]]^{-1},S_0),\\
%&&\\
&&\hspace{-.7cm}\mathrm{where,}\;\;\phi^*=\varphi\land\exists e.\;\mathit{Causes}(\beta(e),t,\varphi).
\end{eqnarray*}
\end{small}
%%}
\end{itemize}
\end{definition}
These can be justified directly using Proposition \ref{CausesProp}, \ref{CAfterProp}, and \ref{CCausesProp} above. 
Intuitively, the additional definitions in the extended regression reduce causes in scenario $do(a,s)$ to that in scenario $s$ via reasoning by cases and using the definition of causes and regression;\footnote{Notice that the formula inside the context of $\mathit{Causes}$ is situation-suppressed. On the other hand, the regression operator $\mathcal{R}_\mathit{ext}$ requires a situation argument. To deal with this, here we  introduce a new situation variable $s'$ and use the $\varphi^{-1}$ operator from \cite{SchLev03} that suppresses the situation argument of $\varphi$ by removing the last (situation) argument of all the fluents in $\varphi$. Thus, e.g., $\mathcal{R}_\mathit{ext}[\varphi[do(\alpha_n(e),s')]]^{-1}$ introduces the situation term $do(\alpha_n(e),s')$ to the situation suppressed formula $\varphi$, performs the regression, and then suppresses the situation argument in the result.}
%of $\varphi$.} 
%
reduce certainly/possibly causes relative to a sequence to that for a shorter sequence, and eventually to a regressable situation calculus formula, which is then regressed;\footnote{Recall that $\mathit{CCauses}$ and $\mathit{PCauses}$ do not take a situation argument, but a sequence of agent actions, which is assumed to be executed starting in $S_0$. We could have generalized this, however.} 
and reduce certainly/possibly causes relative to a (non-empty) sequence of agent actions to a certainly/possibly after formula relative to a shorter sub-sequence, which can then be further reduced using case (5) above.
\par
Having defined this extended regression operator, we are now ready to present our key result:
%{\color{red}
\begin{theorem}\label{Regression theorem}
If $\phi$ is an extended regressable formula and $\mathcal{D}$ is an NDBAT, then:
\begin{eqnarray*}
&& \mathcal{D}\models \phi~~\text{iff}~~\mathcal{D}_{S_0}\cup\mathcal{D}_{una}\models \mathcal{R}^*_{ext}[\phi].
\end{eqnarray*}
\end{theorem}
%%}
%{\color{blue}[Proof sketch?]}
{\color{CBRed}The proof of this theorem is similar to that of the regression theorem in the SC \cite{PirRei99,Reiter01}, but uses Propositions \ref{CausesProp} to \ref{CCausesProp} for the additional cases.
}
%
%\newpage
%{\color{blue}
%\textbf{[I worked on the paper up to this point today; had many issues, including a missing example section! I plan to work on the following example and the conclusion tomorrow (Tuesday)]-Shakil.}
%}
\paragraph{Example (Cont'd).} Consider the agent action sequence $\vec{\alpha_2}=\mathit{move}(I_0,I_1);$ $\mathit{move}(I_1,I_2)$. %, where the robot moves from location $I_0$ to $I_1$ and then from $I_1$ to $I_2$, starting in $S_0$. 
Note that, %moving from $I_1$ to $I_2$ is a certainly cause of vulnerability after executing $\vec{\alpha_2}$, i.e. 
$\mathcal{D}_1\models\mathit{CCauses}(\mathit{move}(I_0, I_1), 0, \mathit{Vul}, \vec{\alpha_2}, S_0)$. We can show that: %this can be converted into the following first order formulas.
%
%{\color{red}
\begin{proposition}
\begin{small}
\begin{eqnarray*}
&&\hspace{-1.1cm}\mathcal{R}_{ext}[\mathit{CCauses}(
\mathit{move}(I_0, I_1), 0, \mathit{Vul}, \vec{\alpha_2})]=\\
%&& \hspace{-.7 cm}\mathit{CAfter}([\mathit{move}(I_0, I_1)],\\
%&&\forall e'.\;\mathit{Poss}(\mathit{move}(I_1,I_2,e'))\\
%&&\hspace{7 mm}\supset\mathcal{R}_{ext}[\phi^*[do(\mathit{move}(I_1,I_2,e'),s')]]^{-1},S_0),\\
%%&&\\
%&&\hspace{-.7cm}\mathrm{where,}\;\;\phi^*=\neg\mathit{Vul}\vee\exists e.\;\mathit{Causes}(\mathit{move}(I_0,I_1,e),t,\mathit{Vul}),
%\\[1ex]
%%&& \equiv \mbox{show the above with } \mathcal{R}_{ext}[\phi^*[do(move(I_1,I_2,e'),s')]]^{-1},S_0) \mbox{ expanded}
&&\hspace{-11 mm}\mathit{CAfter}([\mathit{move}(I_0, I_1)],\\
&&\hspace{-7 mm}\forall e'.\;\mathit{Poss}(\mathit{move}(I_1,I_2,e'))\supset{}\\
&&\hspace{-7 mm}[\neg(e'=\mathit{Vul}\vee\mathit{Vul})\vee{}\\
&&\hspace{-7 mm}(\exists e.\;(\mathit{time}> t \land \mathit{Vul}\\
&&\hspace{2 mm}{}\land 
\mathit{Causes}(\mathit{move}(I_0,I_1,e),t,\mathit{Vul}))\vee{}\\
&&\hspace{-0.3 mm}(\mathit{time} > t\wedge\lnot \mathit{Vul}\land 
e'=\mathit{Vul}\\
&&\hspace{2 mm}{}\land\mathit{Causes}(\mathit{move}(I_0,I_1,e),t,\phi')))],S_0),\\
&&\hspace{-7 mm}\mathrm{where,}\;\;\phi'=\mathit{Poss}(\mathit{move}(I_1,I_2,e')) \land(e'=\mathit{Vul}\vee\mathit{Vul}).
\end{eqnarray*}
%%}
\end{small}
\end{proposition}
Repeated applications of $\mathcal{R}_\mathit{ext}$ yields a formula about $S_0$ that can be 
checked against 
%verified to be true using 
$\mathcal{D}_{S_0}\cup\mathcal{D}_\mathit{una}$ in $\mathcal{D}_1$. 
{\color{CBRed}The complete derivation is shown in the technical appendix.}
\section{Conclusion}
Motivated by the nondeterministic nature of real world action and change, in this paper, we studied reasoning about actual achievement causes in the NDSC \cite{DeGiacomoLespKR21}. We showed that in such domains, when the environment reactions are not known, two different notions of causes can be defined, one where an agent action is a cause for all possible environment reactions, and thus it is certainly a cause, and another where the agent action is a cause for at least one environment choice, i.e. it is possibly a cause. Extending on previous work, we also showed that one can define a regression operator in the situation calculus to reason about these notions. {\color{CBRed} Note that our definition of extended regression enjoys the benefits of Reiter’s regression operator in that it completely sidesteps the second-order part of the theory. Khan and Soutchanski \shortcite{KhanS20} reported a Prolog implementation of Batusov and Soutchanski’s \shortcite{BatusovS18} original proposal, which we think can be extended to deal with the nondeterministic case with some effort.
\par
Here, we focused on actual causation, which assumes that the effect was observed. We thus did not attempt to compute the causes of effects that the agent is unsure about. Such incompleteness must be dealt through the epistemics of causation, where in some possible worlds an agent might consider some effect to be true and some actions to be causes of the effect; in some other possible worlds, she might consider another set of actions to be causes; while in yet other worlds, she might even consider the effect to be false. See \cite{KL21} for an account of this in a deterministic setting. Causal knowledge and its dynamics in the context of nondeterministic actions become even more interesting.%, which we are currently investigating.
\par
Besides the SEM-based framework discussed in the introduction, another framework that is also closely related to our work is the one proposed by Banihashemi, Khan, and Soutchanski \shortcite{BKS22}, where the authors define a weak notion of causation, which is intuitively similar to our notion of possibly causes. However, in that paper they primarily focused on abstraction. Also, their semantics is based on the (ordinary) situation calculus, and nondeterminism is handled using nondeterministic ConGolog programs \cite{GiacomoLL00}. This makes the semantics quite complicated, in contrast to our simple proposal. They also did not define a notion of certainly causes or address how reasoning about causes can be performed.
}
\par
As discussed above, perhaps the closest to our work is the preliminary study proposed recently in \cite{RKLY24}, where the authors also looked at causation in nondeterministic domains. However, as mentioned before, reasoning about causes in that framework is defined using a process of direct translation that in practice generates large formulae. In contrast, we showed how one can extend regression in the situation calculus to produce more compact formulae. 
\par
{\color{CBRed}
Much work remains. A key restriction in these recent situation calculus-based action-theoretic proposals (including ours) is that these assume that the scenario is simply a sequence of actions, and thus do not allow concurrent actions; we are currently working to address this.} We are also looking into the epistemics of causation in nondeterministic domains, extending previous work by Khan and Lesp\'{e}rance \shortcite{KL21}.
%\newpage\input{temp}\newpage
{\color{CBRed}
\section*{Acknowledgements}
This work is partially supported by the National Science and Engineering Research Council of Canada, 
by the University of Regina, and by York University.
}
\bibliography{references}

\begin{thebibliography}{30}
\providecommand{\natexlab}[1]{#1}

\bibitem[{Banihashemi, Khan, and Soutchanski(2022)}]{BKS22}
Banihashemi, B.; Khan, S.~M.; and Soutchanski, M. 2022.
\newblock From Actions to Programs as Abstract Actual Causes.
\newblock In \emph{Thirty-Sixth {AAAI} Conference on Artificial Intelligence,
  {AAAI} 2022, Thirty-Fourth Conference on Innovative Applications of
  Artificial Intelligence, {IAAI} 2022, The Twelveth Symposium on Educational
  Advances in Artificial Intelligence, {EAAI} 2022 Virtual Event, February 22 -
  March 1, 2022}, 5470--5478. {AAAI} Press.

\bibitem[{Batusov and Soutchanski(2017)}]{BatusovS17}
Batusov, V.; and Soutchanski, M. 2017.
\newblock Situation Calculus Semantics for Actual Causality.
\newblock In Gordon, A.~S.; Miller, R.; and Tur{\'{a}}n, G., eds.,
  \emph{Proceedings of the Thirteenth International Symposium on Commonsense
  Reasoning, {COMMONSENSE} 2017, London, UK, November 6-8, 2017}, volume 2052
  of \emph{{CEUR} Workshop Proceedings}. CEUR-WS.org.

\bibitem[{Batusov and Soutchanski(2018)}]{BatusovS18}
Batusov, V.; and Soutchanski, M. 2018.
\newblock Situation Calculus Semantics for Actual Causality.
\newblock In McIlraith, S.~A.; and Weinberger, K.~Q., eds., \emph{Proceedings
  of the Thirty-Second {AAAI} Conference on Artificial Intelligence, (AAAI-18),
  the 30th innovative Applications of Artificial Intelligence (IAAI-18), and
  the 8th {AAAI} Symposium on Educational Advances in Artificial Intelligence
  (EAAI-18), New Orleans, Louisiana, USA, February 2-7, 2018}, 1744--1752.
  {AAAI} Press.

\bibitem[{Beckers(2024)}]{SB24}
Beckers, S. 2024.
\newblock Nondeterministic Causal Models.
\newblock \emph{CoRR}, abs/2405.14001.

\bibitem[{Beckers and Vennekens(2018)}]{BeckersV18}
Beckers, S.; and Vennekens, J. 2018.
\newblock A Principled Approach to Defining Actual Causation.
\newblock \emph{Synthese}, 195(2): 835--862.

\bibitem[{Bochman(2018)}]{Bochman18}
Bochman, A. 2018.
\newblock Actual Causality in a Logical Setting.
\newblock In Lang, J., ed., \emph{Proceedings of the Twenty-Seventh
  International Joint Conference on Artificial Intelligence, {IJCAI} 2018, July
  13-19, 2018, Stockholm, Sweden}, 1730--1736. ijcai.org.

\bibitem[{{De Giacomo} and Lesp{\'{e}}rance(2021)}]{DeGiacomoLespKR21}
{De Giacomo}, G.; and Lesp{\'{e}}rance, Y. 2021.
\newblock The Nondeterministic Situation Calculus.
\newblock In Bienvenu, M.; Lakemeyer, G.; and Erdem, E., eds.,
  \emph{Proceedings of the 18th International Conference on Principles of
  Knowledge Representation and Reasoning, {KR} 2021, Online event, November
  3-12, 2021}, 216--226.

\bibitem[{de~Lima and Lorini(2024)}]{LimaL24}
de~Lima, T.; and Lorini, E. 2024.
\newblock Model Checking Causality.
\newblock In \emph{Proceedings of the Thirty-Third International Joint
  Conference on Artificial Intelligence, {IJCAI} 2024, Jeju, South Korea,
  August 3-9, 2024}, 3324--3332. ijcai.org.

\bibitem[{Eiter and Lukasiewicz(2002)}]{EiterL02}
Eiter, T.; and Lukasiewicz, T. 2002.
\newblock Complexity Results for Structure-based Causality.
\newblock \emph{Artificial Intelligence}, 142(1): 53--89.

\bibitem[{Giacomo, Lesp{\'{e}}rance, and Levesque(2000)}]{GiacomoLL00}
Giacomo, G.~D.; Lesp{\'{e}}rance, Y.; and Levesque, H.~J. 2000.
\newblock ConGolog, A Concurrent Programming Language based on the Situation
  Calculus.
\newblock \emph{Artificial Intelligence}, 121(1-2): 109--169.

\bibitem[{Glymour et~al.(2010)Glymour, Danks, Glymour, Eberhardt, Ramsey,
  Scheines, Spirtes, Teng, and Zhang}]{GlymourDGERSSTZ10}
Glymour, C.; Danks, D.; Glymour, B.; Eberhardt, F.; Ramsey, J.~D.; Scheines,
  R.; Spirtes, P.; Teng, C.~M.; and Zhang, J. 2010.
\newblock Actual Causation: A Stone Soup Essay.
\newblock \emph{Synthese}, 175(2): 169--192.

\bibitem[{Halpern(2000)}]{Halpern00}
Halpern, J.~Y. 2000.
\newblock Axiomatizing Causal Reasoning.
\newblock \emph{Journal of Artificial Intelligence Research}, 12: 317--337.

\bibitem[{Halpern(2015)}]{Halpern15}
Halpern, J.~Y. 2015.
\newblock A Modification of the Halpern-Pearl Definition of Causality.
\newblock In Yang, Q.; and Wooldridge, M.~J., eds., \emph{Proceedings of the
  Twenty-Fourth International Joint Conference on Artificial Intelligence,
  {IJCAI} 2015, Buenos Aires, Argentina, July 25-31, 2015}, 3022--3033. {AAAI}
  Press.

\bibitem[{Halpern(2016)}]{Halpern16}
Halpern, J.~Y. 2016.
\newblock \emph{Actual Causality}.
\newblock {MIT} Press.
\newblock ISBN 978-0-262-03502-6.

\bibitem[{Halpern and Pearl(2005)}]{HalpernP05}
Halpern, J.~Y.; and Pearl, J. 2005.
\newblock Causes and Explanations: A Structural-Model Approach. Part I: Causes.
\newblock \emph{The British Journal for the Philosophy of Science}, 56(4):
  843--887.

\bibitem[{Hopkins(2005)}]{Hopkins05}
Hopkins, M. 2005.
\newblock \emph{The Actual Cause: From Intuition to Automation}.
\newblock Ph.D. thesis, University of California Los Angeles.

\bibitem[{Hopkins and Pearl(2007)}]{HopkinsP07}
Hopkins, M.; and Pearl, J. 2007.
\newblock Causality and Counterfactuals in the Situation Calculus.
\newblock \emph{Journal of Logic and Computation}, 17(5): 939--953.

\bibitem[{Khan and Lesp{\'{e}}rance(2021)}]{KL21}
Khan, S.~M.; and Lesp{\'{e}}rance, Y. 2021.
\newblock Knowing Why - On the Dynamics of Knowledge about Actual Causes in the
  Situation Calculus.
\newblock In Dignum, F.; Lomuscio, A.; Endriss, U.; and Now{\'{e}}, A., eds.,
  \emph{{AAMAS} `21: 20th International Conference on Autonomous Agents and
  Multiagent Systems, Virtual Event, United Kingdom, May 3-7, 2021}, 701--709.
  {ACM}.

\bibitem[{Khan and Rostamigiv(2023)}]{KR23}
Khan, S.~M.; and Rostamigiv, M. 2023.
\newblock On Explaining Agent Behaviour via Root Cause Analysis: {A} Formal
  Account Grounded in Theory of Mind.
\newblock In Gal, K.; Now{\'{e}}, A.; Nalepa, G.~J.; Fairstein, R.; and
  Radulescu, R., eds., \emph{{ECAI} 2023 - 26th European Conference on
  Artificial Intelligence, September 30 - October 4, 2023, Krak{\'{o}}w,
  Poland}, volume 372 of \emph{Frontiers in Artificial Intelligence and
  Applications}, 1239--1247. {IOS} Press.

\bibitem[{Khan and Soutchanski(2020)}]{KhanS20}
Khan, S.~M.; and Soutchanski, M. 2020.
\newblock Necessary and Sufficient Conditions for Actual Root Causes.
\newblock In Giacomo, G.~D.; Catal{\'{a}}, A.; Dilkina, B.; Milano, M.; Barro,
  S.; Bugar{\'{\i}}n, A.; and Lang, J., eds., \emph{{ECAI} 2020 - 24th European
  Conference on Artificial Intelligence, 29 August-8 September 2020, Santiago
  de Compostela, Spain, August 29 - September 8, 2020 - Including 10th
  Conference on Prestigious Applications of Artificial Intelligence {(PAIS}
  2020)}, volume 325 of \emph{Frontiers in Artificial Intelligence and
  Applications}, 800--808. {IOS} Press.

\bibitem[{Leitner{-}Fischer and Leue(2013)}]{Leitner-FischerL13}
Leitner{-}Fischer, F.; and Leue, S. 2013.
\newblock Causality Checking for Complex System Models.
\newblock In Giacobazzi, R.; Berdine, J.; and Mastroeni, I., eds.,
  \emph{Verification, Model Checking, and Abstract Interpretation, 14th
  International Conference, {VMCAI} 2013, Rome, Italy, January 20-22, 2013.
  Proceedings}, volume 7737 of \emph{Lecture Notes in Computer Science},
  248--267. Springer.

\bibitem[{Levesque, Pirri, and Reiter(1998)}]{LevPirRei98}
Levesque, H.~J.; Pirri, F.; and Reiter, R. 1998.
\newblock Foundations for the Situation Calculus.
\newblock \emph{Electronic Transactions on Artificial Intelligence (ETAI)}, 2:
  159--178.

\bibitem[{McCarthy and Hayes(1969)}]{McCarthyH69}
McCarthy, J.; and Hayes, P.~J. 1969.
\newblock Some Philosophical Problems from the Standpoint of Artificial
  Intelligence.
\newblock \emph{Machine Intelligence}, 4: 463--502.

\bibitem[{Pearl(1998)}]{Pearl98}
Pearl, J. 1998.
\newblock On the Definition of Actual Cause.
\newblock Technical Report R-259, University of California Los Angeles.

\bibitem[{Pearl(2000)}]{Pearl00}
Pearl, J. 2000.
\newblock \emph{Causality: Models, Reasoning, and Inference}.
\newblock Cambridge University Press.

\bibitem[{Pirri and Reiter(1999)}]{PirRei99}
Pirri, F.; and Reiter, R. 1999.
\newblock Some contributions to the metatheory of the situation calculus.
\newblock \emph{J. ACM}, 46(3): 325--361.

\bibitem[{Reiter(2001)}]{Reiter01}
Reiter, R. 2001.
\newblock \emph{Knowledge in Action. Logical Foundations for Specifying and
  Implementing Dynamical Systems}.
\newblock Cambridge, MA, USA: MIT Press.
\newblock ISBN 9780262182188.

\bibitem[{Rostamigiv et~al.(2024)Rostamigiv, Khan, Lesp{\'{e}}rance, and
  Yadkoo}]{RKLY24}
Rostamigiv, M.; Khan, S.~M.; Lesp{\'{e}}rance, Y.; and Yadkoo, M. 2024.
\newblock A Logic of Actual Cause for Non-Deterministic Dynamic Domains.
\newblock In \emph{Working Notes of the 21st European Conference on Multi-Agent
  Systems, August 26-28th, 2024, Dublin, Ireland}.

\bibitem[{Scherl and Levesque(2003)}]{SchLev03}
Scherl, R.~B.; and Levesque, H.~J. 2003.
\newblock Knowledge, action, and the frame problem.
\newblock \emph{Artificial Intelligence}, 144(1-2): 1--39.

\bibitem[{Yazdanpanah et~al.(2023)Yazdanpanah, Gerding, Stein, Dastani, Jonker,
  Norman, and Ramchurn}]{YazdanpanahGSDJNR23}
Yazdanpanah, V.; Gerding, E.~H.; Stein, S.; Dastani, M.; Jonker, C.~M.; Norman,
  T.~J.; and Ramchurn, S.~D. 2023.
\newblock Reasoning about responsibility in autonomous systems: challenges and
  opportunities.
\newblock \emph{{AI} Soc.}, 38(4): 1453--1464.

\end{thebibliography}
%\newpage\input{v1/RC}\newpage
\newpage\section{Technical Appendix}
\subsection{First Application of $\mathcal{R}_\mathit{ext}$}
In the following, we show the full derivation of the application of the $\mathcal{R}_\mathit{ext}$ operator in Proposition 6. First, we apply case (6) of $\mathcal{R}_\mathit{cal}$ to obtain the following.
%{\color{red}
\par
%\noindent\textbf{Proposition 6}
\begin{small}
\begin{eqnarray*}
&&\hspace{-.7cm}\mathcal{R}_{ext}[\mathit{CCauses}(
\mathit{move}(I_0, I_1), 0, \mathit{Vul}, \vec{\alpha_2})]=\\
&& \hspace{-.3 cm}\mathit{CAfter}([\mathit{move}(I_0, I_1)],\\
&&\forall e'.\;\mathit{Poss}(\mathit{move}(I_1,I_2,e'))\\
&&\hspace{7 mm}\supset\mathcal{R}_{ext}[\phi^*[do(\mathit{move}(I_1,I_2,e'),s')]]^{-1},S_0),\\
%&&\\
&&\hspace{-.7cm}\mathrm{where,}\;\;\phi^*=\neg\mathit{Vul}\vee\exists e.\;\mathit{Causes}(\mathit{move}(I_0,I_1,e),0,\mathit{Vul}).
%\\[1ex]
%&& \equiv \mbox{show the above with } \mathcal{R}_{ext}[\phi^*[do(move(I_1,I_2,e'),s')]]^{-1},S_0) \mbox{ expanded}
%
%
%&&\\
\end{eqnarray*}
\end{small}
\par\noindent
Applying case (3) of $\mathcal{R}_\mathit{ext}$, the right-hand side of the above $\supset$ can be reduced to:
\par
\begin{small}
\begin{eqnarray*}
&&\hspace{-7 mm}\equiv\dots\supset
[\neg\mathcal{R}_\mathit{ext}[\mathit{Vul}[do(\mathit{move}(I_1,I_2,e'),s')]]\vee{}\\
&&\hspace{-7 mm}\exists e.\;\mathcal{R}_\mathit{ext}[\mathit{Causes}(\mathit{move}(I_0,I_1,e),0,\mathit{Vul},do(\mathit{move}(I_1,I_2,e'),s'))]\\
&&\hspace{5 mm}]^{-1},S_0),
%
%
%&&\\
\end{eqnarray*}
\end{small}
\par\noindent
Again, if we apply case (2a) on the first disjunct and case (4) on the second, we have:
\begin{small}
\begin{eqnarray*}
&&\hspace{-7 mm}\equiv\dots\supset{}\\
&&\hspace{-7 mm}[\neg(\exists i,j.\;\mathit{move}(I_1,I_2,e')=\mathit{move}(i,j,\mathit{Vul})\vee\mathit{Vul}[s'])\vee{}\\
&&\hspace{-7 mm}(\exists e.\;(\mathit{time}(s') = 0 \land\mathit{move}(I_0,I_1,e) =\mathit{move}(I_1,I_2,e')\\
&&\hspace{2 mm}{}\land\lnot \mathit{Vul}[s'] \land\mathcal{R}_{ext}[\mathit{Vul}[do(\mathit{move}(I_1,I_2,e'),s')]])\vee{}\\
&&\hspace{-0.3 mm}(\mathit{time}(s') > 0 \land \mathit{Vul}[s'] \land \mathcal{R}_{ext}[\mathit{Vul}[do(\mathit{move}(I_1,I_2,e'),s')]]\\ 
&&\hspace{2 mm}{}\land 
\mathit{Causes}(\mathit{move}(I_0,I_1,e),0,\mathit{Vul},s'))\vee{}\\
&&\hspace{-0.3 mm}(\mathit{time}(s') > 0\wedge\lnot \mathit{Vul}[s'] \land \mathcal{R}_{ext}[\mathit{Vul}[do(\mathit{move}(I_1,I_2,e'),s')]]\\
&&\hspace{2 mm}{}\land\mathit{Causes}(\mathit{move}(I_0,I_1,e),0,\mathit{Poss}(\mathit{move}(I_1,I_2,e')) \land{}\\
&&\hspace{10 mm}\mathcal{R}_{ext}[\mathit{Vul}[do(\mathit{move}(I_1,I_2,e'),s^*)]]^{-1},s')))]^{-1},S_0).
\end{eqnarray*}
\end{small}
\par\noindent
We apply case (2a) again and simplify using unique-names axioms for the action $\mathit{move}$ (that says that $\mathit{move}(i,j,e)=\mathit{move}(i',j',e')\supset i=i'\land j=j'\land e=e'$) to obtain the following.
\par\noindent
\begin{small}
\begin{eqnarray*}
&&\hspace{-7 mm}\equiv\dots\supset{}\\
&&\hspace{-7 mm}[\neg(\exists i,j.\;\mathit{move}(I_1,I_2,e')=\mathit{move}(i,j,\mathit{Vul})\vee\mathit{Vul}[s'])\vee{}\\
&&\hspace{-7 mm}(\exists e.\;(
\mathit{false}\vee{}\\
&&\hspace{-0.3 mm}(\mathit{time}(s') > 0 \land \mathit{Vul}[s']\\
&&\hspace{2 mm}{}\land(\exists i,j.\;\mathit{move}(I_1,I_2,e')=\mathit{move}(i,j,\mathit{Vul})\vee\mathit{Vul}[s'])\\ 
&&\hspace{2 mm}{}\land 
\mathit{Causes}(\mathit{move}(I_0,I_1,e),0,\mathit{Vul},s'))\vee{}\\
&&\hspace{-0.3 mm}(\mathit{time}(s') > 0\wedge\lnot \mathit{Vul}[s']\\
&&\hspace{2 mm}{}\land 
(\exists i,j.\;\mathit{move}(I_1,I_2,e')=\mathit{move}(i,j,\mathit{Vul})\vee\mathit{Vul}[s'])\\
&&\hspace{2 mm}{}\land\mathit{Causes}(\mathit{move}(I_0,I_1,e),0,\mathit{Poss}(\mathit{move}(I_1,I_2,e'))\land{}\\
&&\hspace{9 mm}[(\exists i,j.\;\mathit{move}(I_1,I_2,e')=\mathit{move}(i,j,\mathit{Vul})\vee\mathit{Vul}[s^*])]^{-1},\\
&&\hspace{-3 mm}s')))]^{-1},S_0).
\end{eqnarray*}
\end{small}
\par\noindent
After some simplification and removing the situation placeholder $s^*$ and $s'$ by applying the $[\cdot]^{-1}$ operator twice, we finally obtain the following for the right-hand side.
\begin{comment}
\begin{small}
\begin{eqnarray*}
&&\equiv\dots\supset{}\\
&&[\neg(e'=\mathit{Vul}\vee\mathit{Vul}[s'])\vee{}\\
&&(\exists e.\;(\mathit{time}(s') > t \land \mathit{Vul}[s'] \land 
\mathit{Causes}(\mathit{move}(I_0,I_1,e),t,\mathit{Vul},s'))\vee{}\\
&&\hspace{6.7 mm}(\mathit{time}(s') > t\wedge\lnot \mathit{Vul}[s'] \land 
e'=\mathit{Vul}\\
&&\hspace{11 mm}{}\land\mathit{Causes}(\mathit{move}(I_0,I_1,e),t,\mathit{Poss}(\mathit{move}(I_1,I_2,e')) \land
(e'=\mathit{Vul}\vee\mathit{Vul}),s')))]^{-1},S_0).
%
%
\end{eqnarray*}
\end{small}
\end{comment}
%
\begin{small}
\begin{eqnarray*}
&&\hspace{-7 mm}\equiv\dots\supset{}\\
&&\hspace{-7 mm}[\neg(e'=\mathit{Vul}\vee\mathit{Vul})\vee{}\\
&&\hspace{-7 mm}(\exists e.\;(\mathit{time} > 0 \land \mathit{Vul} \land 
\mathit{Causes}(\mathit{move}(I_0,I_1,e),0,\mathit{Vul}))\vee{}\\
&&\hspace{-0.3 mm}(\mathit{time} > 0\wedge\lnot \mathit{Vul} \land 
e'=\mathit{Vul}\\
&&\hspace{4 mm}{}\land\mathit{Causes}(\mathit{move}(I_0,I_1,e),0,\\
&&\hspace{11 mm}\mathit{Poss}(\mathit{move}(I_1,I_2,e')) \land(e'=\mathit{Vul}\vee\mathit{Vul}))))\\
&&\hspace{-7 mm}],S_0).
\end{eqnarray*}
%%}
\end{small}
%}
%\end{document}%YYY
%%%%%%%%%%%%%%%%%%%%%%%%%%%%%
%%%%%%%%%%%%%%%%%%%%%%%%%%%%%
%%%%%%%%%%%%%%%%%%%%%%%%%%%%%
%\newpage
\subsection{Second Application of $\mathcal{R}_\mathit{ext}$}
We next show another application of $\mathcal{R}_\mathit{ext}$ on the formula that we obtained above. Below, $\psi$ stands for the formula in square brackets above. We first apply cases (5) and (3) of $\mathcal{R}_\mathit{ext}$ to obtain the following.
\par\noindent
\begin{small}
\begin{eqnarray*}
&& \hspace{-7 mm}\mathcal{R}_{ext}[\mathit{CAfter}([\mathit{move}(I_0, I_1)],\\
&&\hspace{7 mm}\forall e'.\;\mathit{Poss}(\mathit{move}(I_1,I_2,e'))
\supset\psi,S_0)]=\\
&&\hspace{-7 mm}\mathit{CAfter}(\epsilon,\forall e''.\;\mathit{Poss}(\mathit{move}(I_0,I_1,e''))\\
&&\hspace{12 mm}{}\supset[\forall e'.\mathcal{R}_\mathit{ext}[\mathit{Poss}(\mathit{move}(I_1,I_2,e'),s_1)]\\
&&\hspace{19 mm}{}\supset\mathcal{R}_\mathit{ext}[\psi[s_1]]]^{-1},S_0),\\
&&\hspace{-7 mm}\mathrm{where,}\;\;s_1=do(\mathit{move}(I_0,I_1,e''),s').
\end{eqnarray*}
\end{small}
\par\noindent
Let us handle the two regressions above separately, starting with $\mathcal{R}_\mathit{ext}[\mathit{Poss}(\mathit{move}(I_1,I_2,e'),s_1)]^{-1}$, i.e.\ the first one.\footnote{Note, the $[\cdot]^{-1}$ operator is distributive.} 
\par
Applying cases (2c), (3), and (1), we have:
\par\noindent
\begin{small}
\begin{eqnarray*}
&&\hspace{-7 mm}\mathcal{R}_\mathit{ext}[\mathit{Poss}(\mathit{move}(I_1,I_2,e'),s_1)]^{-1}\\
&&\hspace{-7 mm}{}=[\mathcal{R}_\mathit{ext}[\mathit{Poss}_\mathit{ag}(move(I_1,I_2),s_1)]\\
&&\hspace{-3 mm}\mbox{}\land(\mathcal{R}_\mathit{ext}[\mathit{Risky}(I_2,s_1)]\supset(e'=\mathit{Vul} \vee e'=\mathit{NotVul}))\\
&&\hspace{-3 mm}{}\land
(\neg\mathcal{R}_\mathit{ext}[\mathit{Risky}(I_2,s_1)]\supset e'=\mathit{NotVul})]^{-1}\\
&&\hspace{-7 mm}{}\equiv[\mathcal{R}_\mathit{ext}[\mathit{At}(I_1,s_1)]\land\mathit{Connected}(I_1,I_2)\\
&&\hspace{-3 mm}\mbox{}\land(\mathcal{R}_\mathit{ext}[\mathit{Risky}(I_2,s_1)]\supset(e'=\mathit{Vul} \vee e'=\mathit{NotVul}))\\
&&\hspace{-3 mm}{}\land
(\neg\mathcal{R}_\mathit{ext}[\mathit{Risky}(I_2,s_1)]\supset e'=\mathit{NotVul})]^{-1}.
\end{eqnarray*}
\end{small}
\par\noindent
Again, if we apply case (2a), simplify using unique-names axioms for actions and the fact that $I_1$ and $I_2$ are connected (see Fig.\ 1), and apply the $[\cdot]^{-1}$ operator, we obtain:
\par\noindent
\begin{small}
\begin{eqnarray*}
%&&\hspace{-7 mm}\mathcal{R}_\mathit{ext}[\mathit{Poss}(\mathit{move}(I_1,I_2,e'),s_1)]^{-1}\\
%
&&\hspace{-7 mm}{}\equiv[(\exists i_1,e_1.\;\mathit{move}(I_0,I_1,e'')=\mathit{move}(i_1,I_1,e_1))\vee{}\\
&&\hspace{-1.5 mm}(\mathit{At}(I_1,s')\land\forall j_2,e_2.\;\neg (\mathit{move}(I_0,I_1,e'')=\mathit{move}(I_1,j_2,e_2)))]\\
&&\hspace{-3 mm}\mbox{}\land(\mathit{Risky}(I_2,s')\supset(e'=\mathit{Vul} \vee e'=\mathit{NotVul}))\\
&&\hspace{-3 mm}{}\land
(\neg\mathit{Risky}(I_2,s')\supset e'=\mathit{NotVul})]^{-1}\\
&&\hspace{-7 mm}{}\equiv \{ \mathit{At}(I_0) \land (\mathit{Risky}(I_2)\supset(e'=\mathit{Vul} \vee e'=\mathit{NotVul}))\\
&&\hspace{-1 mm}{}\land
(\neg\mathit{Risky}(I_2)\supset e'=\mathit{NotVul})\}.
\end{eqnarray*}
\end{small}
\par\noindent
Next we expand $\mathcal{R}_\mathit{ext}[\psi[s_1]]]^{-1}$. If we apply cases (1) and (3), we have the following.
\par\noindent
\begin{small}
\begin{eqnarray*}
&&\hspace{-7 mm}\mathcal{R}_\mathit{ext}[\psi[s_1]]]^{-1}=\\
&&\hspace{-7 mm}[\neg(e'=\mathit{Vul}\vee\mathcal{R}_\mathit{ext}[\mathit{Vul}[s_1]])\vee{}\\
&&\hspace{-7 mm}(\exists e.\;(\mathcal{R}_\mathit{ext}[\mathit{time}(s_1) > 0] \land \mathcal{R}_\mathit{ext}[\mathit{Vul}[s_1]]\\
&&\hspace{4 mm}{}\land 
\mathcal{R}_\mathit{ext}[\mathit{Causes}(\mathit{move}(I_0,I_1,e),0,\mathit{Vul},s_1)])\vee{}\\
&&\hspace{-0.3 mm}(\mathcal{R}_\mathit{ext}[\mathit{time}(s_1) > 0]\wedge\lnot \mathcal{R}_\mathit{ext}[\mathit{Vul}[s_1]] \land 
e'=\mathit{Vul}\\
&&\hspace{4 mm}{}\land\mathcal{R}_\mathit{ext}[\mathit{Causes}(\mathit{move}(I_0,I_1,e),0,\\
&&\hspace{11 mm}\mathit{Poss}(\mathit{move}(I_1,I_2,e')) \land(e'=\mathit{Vul}\vee\mathit{Vul}),s_1)]))\\
&&\hspace{-7 mm}]^{-1}.
\end{eqnarray*}
\end{small}
%%}
%
\par\noindent
Next we apply case (2a) and (2b) (and the SSA for $\mathit{time}$ and $\mathit{Vul}$), and simplify as before to obtain the following.
\par\noindent
\begin{small}
\begin{eqnarray*}
&&\hspace{-7 mm}{}\equiv[\neg(e'=\mathit{Vul}\vee(e''=\mathit{Vul}\vee\mathit{Vul}[s']))\vee{}\\
&&\hspace{-7 mm}(\exists e.\;(\mathit{time}(s') > -1 \land(e''=\mathit{Vul}\vee\mathit{Vul}[s']) \\
&&\hspace{4 mm}{}\land 
\mathcal{R}_\mathit{ext}[\mathit{Causes}(\mathit{move}(I_0,I_1,e),0,\mathit{Vul},s_1)])\vee{}\\
&&\hspace{-0.3 mm}(\mathit{time}(s') > -1\wedge\lnot (e''=\mathit{Vul}\vee\mathit{Vul}[s']) \land 
e'=\mathit{Vul}\\
&&\hspace{4 mm}{}\land\mathcal{R}_\mathit{ext}[\mathit{Causes}(\mathit{move}(I_0,I_1,e),0,\\
&&\hspace{11 mm}\mathit{Poss}(\mathit{move}(I_1,I_2,e')) \land(e'=\mathit{Vul}\vee\mathit{Vul}),s_1)]))\\
&&\hspace{-7 mm}]^{-1}.
\end{eqnarray*}
\end{small}
%%}
%
\par\noindent
Let us now simplify the two $\mathcal{R}_\mathit{ext}$ above separately, starting with the first one. Using case (4), we have:
\par\noindent
\begin{small}
\begin{eqnarray*}
&&\hspace{-7 mm}\mathcal{R}_\mathit{ext}[\mathit{Causes}(\mathit{move}(I_0,I_1,e),0,\mathit{Vul},s_1)]={}\\
&&\hspace{-2 mm}[(\mathit{time}(s') = 0 \land\mathit{move}(I_0,I_1,e) =\mathit{move}(I_0,I_1,e'')\\
&&\hspace{2 mm}{}\land\lnot \mathit{Vul}[s'] \land\mathcal{R}_{ext}[\mathit{Vul}[do(\mathit{move}(I_0,I_1,e''),s')]])\vee{}\\
&&\hspace{-0.3 mm}(\mathit{time}(s') > 0 \land \mathit{Vul}[s'] \land \mathcal{R}_{ext}[\mathit{Vul}[do(\mathit{move}(I_0,I_1,e''),s')]]\\ 
&&\hspace{2 mm}{}\land 
\mathit{Causes}(\mathit{move}(I_0,I_1,e),0,\mathit{Vul},s'))\vee{}\\
&&\hspace{-0.3 mm}(\mathit{time}(s') > 0\wedge\lnot \mathit{Vul}[s'] \land \mathcal{R}_{ext}[\mathit{Vul}[do(\mathit{move}(I_0,I_1,e''),s')]]\\
&&\hspace{2 mm}{}\land\mathit{Causes}(\mathit{move}(I_0,I_1,e),0,\mathit{Poss}(\mathit{move}(I_0,I_1,e'')) \land{}\\
&&\hspace{10 mm}\mathcal{R}_{ext}[\mathit{Vul}[do(\mathit{move}(I_0,I_1,e''),s^*)]]^{-1},s'))].
\end{eqnarray*}
\end{small}
\par\noindent
If we apply case (2a), operator $[\cdot]^{-1}$ once, and simplify, we have:
\par\noindent
\begin{small}
\begin{eqnarray*}
%&&\hspace{-7 mm}\mathcal{R}_\mathit{ext}[\mathit{Causes}(\mathit{move}(I_0,I_1,e),t,\mathit{Vul},s_1)]={}\\
%
&&\hspace{-5.5 mm}{}\equiv[(\mathit{time}(s') = 0 \land e = e''\land\lnot \mathit{Vul}[s']\\
&&\hspace{2 mm}{} \land(e''=\mathit{Vul}\vee\mathit{Vul}[s']))\vee{}\\
&&\hspace{-0.3 mm}(\mathit{time}(s') > 0 \land \mathit{Vul}[s']\land(e''=\mathit{Vul}\vee\mathit{Vul}[s'])\\ 
&&\hspace{2 mm}{}\land 
\mathit{Causes}(\mathit{move}(I_0,I_1,e),0,\mathit{Vul},s'))\vee{}\\
&&\hspace{-0.3 mm}(\mathit{time}(s') > 0\wedge\lnot \mathit{Vul}[s'] \land (e''=\mathit{Vul}\vee\mathit{Vul}[s'])\\
&&\hspace{2 mm}{}\land\mathit{Causes}(\mathit{move}(I_0,I_1,e),0,\\
&&\hspace{10 mm}\mathit{Poss}(\mathit{move}(I_0,I_1,e'')) \land(e''=\mathit{Vul}\vee\mathit{Vul}),s'))].
\end{eqnarray*}
\end{small}
\par\noindent
Simplifying further, thus we have:
\par\noindent
\begin{small}
\begin{eqnarray*}
&&\hspace{-7 mm}\mathcal{R}_\mathit{ext}[\mathit{Causes}(\mathit{move}(I_0,I_1,e),0,\mathit{Vul},s_1)]={}\\
&&\hspace{-1 mm}[(\mathit{time}(s') = 0 \land e = e''\land\lnot \mathit{Vul}[s']\land e''=\mathit{Vul})\vee{}\\
&&\hspace{-0.3 mm}(\mathit{time}(s') > 0 \land \mathit{Vul}[s']\\ 
&&\hspace{2 mm}{}\land 
\mathit{Causes}(\mathit{move}(I_0,I_1,e),0,\mathit{Vul},s'))\vee{}\\
&&\hspace{-0.3 mm}(\mathit{time}(s') > 0\wedge\lnot \mathit{Vul}[s'] \land e''=\mathit{Vul}\\
&&\hspace{2 mm}{}\land\mathit{Causes}(\mathit{move}(I_0,I_1,e),0,\\
&&\hspace{10 mm}\mathit{Poss}(\mathit{move}(I_0,I_1,e'')) \land(e''=\mathit{Vul}\vee\mathit{Vul}),s'))].
\end{eqnarray*}
\end{small}
\par\noindent
Returning to the expansion of $\mathcal{R}_\mathit{ext}[\psi[s_1]]]^{-1}$, the second $\mathcal{R}_\mathit{ext}(\mathit{Causes}(\dots))$ is similar, with the fluent $\mathit{Vul}$ replaced by $\phi_a$, where $\phi_a=\mathit{Poss}(\mathit{move}(I_1,I_2,e')) \land(e'=\mathit{Vul}\vee\mathit{Vul}).$ Let us compute this now.
\par\noindent
%FIRST
\begin{small}
\begin{eqnarray*}
&&\hspace{-7 mm}\mathcal{R}_\mathit{ext}[\mathit{Causes}(\mathit{move}(I_0,I_1,e),0,\phi_a,s_1)]={}\\
&&\hspace{-2 mm}[(\mathit{time}(s') = 0 \land\mathit{move}(I_0,I_1,e) =\mathit{move}(I_0,I_1,e'')\\
&&\hspace{2 mm}{}\land\lnot\phi_a[s'] \land\mathcal{R}_{ext}[\phi_a[s_1]])\vee{}\\
&&\hspace{-0.3 mm}(\mathit{time}(s') > 0 \land \phi_a[s'] \land \mathcal{R}_{ext}[\phi_a[s_1]]\\ 
&&\hspace{2 mm}{}\land 
\mathit{Causes}(\mathit{move}(I_0,I_1,e),0,\phi_a,s'))\vee{}\\
&&\hspace{-0.3 mm}(\mathit{time}(s') > 0\wedge\lnot \phi_a[s'] \land \mathcal{R}_{ext}[\phi_a[s_1]]\\
&&\hspace{2 mm}{}\land\mathit{Causes}(\mathit{move}(I_0,I_1,e),0,\mathit{Poss}(\mathit{move}(I_0,I_1,e'')) \land{}\\
&&\hspace{10 mm}\mathcal{R}_{ext}[\phi_a[do(\mathit{move}(I_0,I_1,e''),s^*)]]^{-1},s'))].
\end{eqnarray*}
\end{small}
%SECOND
\par\noindent
Let us compute $\mathcal{R}_\mathit{ext}[\phi_a[s_1]]^{-1}$.
\par\noindent
\begin{small}
\begin{eqnarray*}
&&\hspace{-7 mm}\mathcal{R}_\mathit{ext}[\phi_a[s_1]]^{-1}\\
&&\hspace{-7 mm}{}=\mathcal{R}_\mathit{ext}[\mathit{Poss}(\mathit{move}(I_1,I_2,e'),s_1)]^{-1} \land(e'=\mathit{Vul}\vee\mathcal{R}_\mathit{ext}[\mathit{Vul}[s_1]]^{-1})\\
&&\hspace{-7 mm}{}\equiv \mathit{At}(I_0) \land (\mathit{Risky}(I_2)\supset(e'=\mathit{Vul}\vee e'=\mathit{NotVul}))\\
&&
{}\land(\neg\mathit{Risky}(I_2)\supset e'=\mathit{NotVul})\\
&&{}\land(e''=\mathit{Vul}\vee \mathit{Vul})
\end{eqnarray*}
\end{small}
\par\noindent
Thus we have that:
\par\noindent
\begin{small}
\begin{eqnarray*}
&&\hspace{-7 mm}\mathcal{R}_\mathit{ext}[\mathit{Causes}(\mathit{move}(I_0,I_1,e),0,\phi_a,s_1)]={}\\
&&\hspace{-2 mm}[(\mathit{time}(s') = 0 \land e = e'' \land\lnot\phi_a[s']\land\mathcal{R}_{ext}[\phi_a[s_1]])\vee{}\\
&&\hspace{-0.3 mm}(\mathit{time}(s') > 0 \land \phi_a[s'] \land \mathcal{R}_{ext}[\phi_a[s_1]]\\ 
&&\hspace{2 mm}{}\land 
\mathit{Causes}(\mathit{move}(I_0,I_1,e),0,\phi_a,s'))\vee{}\\
&&\hspace{-0.3 mm}(\mathit{time}(s') > 0\wedge\lnot \phi_a[s'] \land \mathcal{R}_{ext}[\phi_a[s_1]]\\
&&\hspace{2 mm}{}\land\mathit{Causes}(\mathit{move}(I_0,I_1,e),0,\mathit{Poss}(\mathit{move}(I_0,I_1,e'')) \land{}\\
&&\hspace{10 mm}\mathcal{R}_{ext}[\phi_a[do(\mathit{move}(I_0,I_1,e''),s^*)]]^{-1},s'))].
\end{eqnarray*}
\end{small}
\par\noindent
If we plug in these two regression results in the right-hand side of the original formula and apply the $[\cdot]^{-1}$ operator, we get:\footnote{For convenience, we give the previous step again.}
\par\noindent
\begin{small}
\begin{eqnarray*}
&&\hspace{-7 mm}\mathcal{R}_\mathit{ext}[\psi[s_1]]]^{-1}\\
&&\hspace{-7 mm}{}=[\neg(e'=\mathit{Vul}\vee(e''=\mathit{Vul}\vee\mathit{Vul}[s']))\vee{}\\
&&\hspace{-7 mm}(\exists e.\;(\mathit{time}(s') > -1 \land(e''=\mathit{Vul}\vee\mathit{Vul}[s']) \\
&&\hspace{4 mm}{}\land 
\mathcal{R}_\mathit{ext}[\mathit{Causes}(\mathit{move}(I_0,I_1,e),0,\mathit{Vul},s_1)])\vee{}\\
&&\hspace{-0.3 mm}(\mathit{time}(s') > -1\wedge\lnot (e''=\mathit{Vul}\vee\mathit{Vul}[s']) \land 
e'=\mathit{Vul}\\
&&\hspace{4 mm}{}\land\mathcal{R}_\mathit{ext}[\mathit{Causes}(\mathit{move}(I_0,I_1,e),0,\\
&&\hspace{11 mm}\mathit{Poss}(\mathit{move}(I_1,I_2,e')) \land(e'=\mathit{Vul}\vee\mathit{Vul}),s_1)]))%\\&&\hspace{-7 mm}
]^{-1}\\[1ex]
% %&&\\
 \end{eqnarray*}
% \end{small}
% %%%%%
% \begin{small}
 \begin{eqnarray*}
&&\hspace{-7 mm}{}\equiv
\{\neg(e'=\mathit{Vul}\vee(e''=\mathit{Vul}\vee\mathit{Vul}))\vee{}\\
&&\hspace{-7 mm}(\exists e.(\mathit{time}> -1 \land(e''=\mathit{Vul}\vee\mathit{Vul}) \\
&&\hspace{1 mm}{}\land[(\mathit{time} = 0 \land e = e''\land\lnot \mathit{Vul}\land e''=\mathit{Vul})\vee{}\\
&&\hspace{5.5 mm}(\mathit{time} > 0 \land \mathit{Vul}%\\ &&\hspace{2 mm}{}
\land 
\mathit{Causes}(\mathit{move}(I_0,I_1,e),0,\mathit{Vul}))\vee{}\\
&&\hspace{5.5 mm}(\mathit{time} > 0\wedge\lnot \mathit{Vul} \land e''=\mathit{Vul}\\
&&\hspace{7.5 mm}{}\land\mathit{Causes}(\mathit{move}(I_0,I_1,e),0,\\
&&\hspace{13 mm}\mathit{Poss}(\mathit{move}(I_0,I_1,e'')) \land(e''=\mathit{Vul}\vee\mathit{Vul})))])\vee{}\\
&&\hspace{-0.3 mm}(\mathit{time}> -1\wedge\lnot (e''=\mathit{Vul}\vee\mathit{Vul}) \land 
e'=\mathit{Vul}\\
%RExt2
&&\hspace{1 mm}{}\land
[(\mathit{time} = 0 \land e = e''\land\lnot \phi_a\land \mathcal{R}_\mathit{ext}[\phi_a[s_1]]^{-1}\vee{}\\
&&\hspace{5.5 mm}(\mathit{time} > 0 \land \phi_a \land \mathcal{R}_\mathit{ext}[\phi_a[s_1]]^{-1}\\ 
&&\hspace{7.5 mm}{} \land 
\mathit{Causes}(\mathit{move}(I_0,I_1,e),0,\phi_a))\vee{}\\
&&\hspace{5.5 mm}(\mathit{time} > 0\wedge\lnot \phi_a \land \mathcal{R}_\mathit{ext}[\phi_a[s_1]]^{-1}\\
&&\hspace{7.5 mm}{}\land\mathit{Causes}(\mathit{move}(I_0,I_1,e),0,\\
&&\hspace{13 mm}\mathit{Poss}(\mathit{move}(I_0,I_1,e'')) \land \mathcal{R}_\mathit{ext}[\phi_a[s_1]]^{-1})]))\\
&&\hspace{-7 mm}\}.
\end{eqnarray*}
\end{small}
\par\noindent
Thus the result of our second application of $\mathcal{R}_\mathit{ext}$ is:
\par\noindent
\begin{small}
\begin{eqnarray*}
&&\hspace{-7 mm}\mathit{CAfter}(\epsilon,\forall e''.\;\mathit{Poss}(\mathit{move}(I_0,I_1,e''))
%\\&&\hspace{15 mm}{}
\supset\forall e'.(\psi_a\supset\psi_b),S_0),
\end{eqnarray*}
\end{small}
\par\noindent
where $\psi_a$ is the formula in the curly braces in the previous page and $\psi_b$ is the above formula in the curly braces.
\par\noindent
%%%%%%%%%%%%%%%%%%%%%%%%%%%%%
%%%%%%%%%%%%%%%%%%%%%%%%%%%%%
%%%%%%%%%%%%%%%%%%%%%%%%%%%%%
\subsection{Third Application of $\mathcal{R}_\mathit{ext}$}
\par\noindent
We now apply $\mathcal{R}_\mathit{ext}$ a third time on the formula obtained above, cases (5) and (3) of which give us the following.
\par\noindent
\begin{small}
\begin{eqnarray*}
&&\hspace{-7 mm}\mathcal{R}_\mathit{ext}[\mathit{CAfter}(\epsilon,\forall e''.\;\mathit{Poss}(\mathit{move}(I_0,I_1,e''))\\
&&\hspace{19 mm}{}\supset\forall e'.(\psi_a\supset\psi_b),S_0)]\\
&&\hspace{-7 mm}=\forall e''.\;\mathcal{R}_\mathit{ext}[\mathit{Poss}(\mathit{move}(I_0,I_1,e''),S_0)]\\
&&\hspace{0 mm}{}\supset\forall e'.(\mathcal{R}_\mathit{ext}[\psi_a[S_0]]\supset\mathcal{R}_\mathit{ext}[\psi_b[S_0]]).
\end{eqnarray*}
\end{small}
\par\noindent
Let us handle the extended regression of the $\mathit{Poss}$, $\psi_a$, and $\psi_b$ separately, starting with the $\mathit{Poss}$.
\par
Now, applying case (2c) and simplifying using initial state axioms (7) and (8) on page 3 and the fact that $\mathit{connected}(I_0,I_1)$ (see Fig.\ 1), we have:
\par\noindent
\begin{small}
\begin{eqnarray*}
&&\hspace{-7 mm}\mathcal{R}_\mathit{ext}[\mathit{Poss}(\mathit{move}(I_0,I_1,e''),S_0)]\\
&&\hspace{-7 mm}{}=[\mathit{Poss}_\mathit{ag}(move(I_0,I_1),S_0)\\
&&\hspace{-3 mm}\mbox{}\land(\mathit{Risky}(I_1,S_0)\supset(e''=\mathit{Vul} \vee e''=\mathit{NotVul}))\\
&&\hspace{-3 mm}{}\land
(\neg\mathit{Risky}(I_1,S_0)\supset e''=\mathit{NotVul})]\\
&&\hspace{-7 mm}{}\equiv[\mathit{At}(I_0,S_0)\land \mathit{Connected}(I_0,I_1)\\
&&\hspace{-3 mm}\mbox{}\land(\mathit{Risky}(I_1,S_0)\supset(e''=\mathit{Vul} \vee e''=\mathit{NotVul}))\\
&&\hspace{-3 mm}{}\land
(\neg\mathit{Risky}(I_1,S_0)\supset e''=\mathit{NotVul})]\\
&&\hspace{-7 mm}{}\equiv\{e''=\mathit{Vul}\vee e''=\mathit{NotVul}\}.
\end{eqnarray*}
\end{small}
\par\noindent
Next, let us compute $\mathcal{R}_\mathit{ext}[\psi_a[S_0]]$. If we apply rules (3) and (1) and simplify using Axiom (8) on page 3, we have:
\par\noindent
\begin{small}
\begin{eqnarray*}
&&\hspace{-7 mm}\mathcal{R}_\mathit{ext}[\psi_a[S_0]]\\
&&\hspace{-7 mm}{}=\mathcal{R}_\mathit{ext}[(\mathit{Risky}(I_2,S_0)\supset(e'=\mathit{Vul} \vee e'=\mathit{NotVul}))\\
&&\hspace{-3 mm}{}\land
(\neg\mathit{Risky}(I_2,S_0)\supset e'=\mathit{NotVul})]\\
&&\hspace{-7 mm}{}\equiv\{e'=\mathit{Vul}\vee e'=\mathit{NotVul}\}.
\end{eqnarray*}
\end{small}
\par
Lastly, let us deal with the $\mathcal{R}_\mathit{ext}[\psi_b[S_0]]$. Note that by definition, $\mathit{Causes}(a,ts,\varphi,S_0)$ is equivalent to $\mathit{false}$ for any $a,ts,\varphi$, so $\mathcal{R}_\mathit{ext}[\mathit{Causes}(a,ts,\varphi,S_0)]=\mathit{false}$. Thus, by cases (3) and (1) and the $\mathit{time}$ axiom that $\mathit{time}(S_0)=0$, we have:
\par\noindent
\begin{small}
\begin{eqnarray*}
&&\hspace{-7 mm}\mathcal{R}_\mathit{ext}[\psi_b[S_0]]\\
&&\\
&&\hspace{-7 mm}{}=\neg(e'=\mathit{Vul}\vee(e''=\mathit{Vul}\vee\mathit{Vul}[S_0]))\vee{}\\
&&\hspace{-7 mm}(\exists e.(\mathit{time}(S_0)> -1 \land(e''=\mathit{Vul}\vee\mathit{Vul}[S_0]) \\
&&\hspace{1 mm}{}\land[(\mathit{time}(S_0) = 0 \land e = e''\land\lnot \mathit{Vul}[S_0]\land e''=\mathit{Vul})\vee{}\\
&&\hspace{5.5 mm}(\mathit{time}(S_0) > 0 \land \mathit{Vul}[S_0]%\\ &&\hspace{2 mm}{}
\land 
\mathit{false})\vee{}\\
&&\hspace{5.5 mm}(\mathit{time}(S_0) > 0\wedge\lnot \mathit{Vul}[S_0] \land e''=\mathit{Vul}\land\mathit{false})])\vee{}\\
&&\hspace{-0.3 mm}(\mathit{time}(S_0)> -1\wedge\lnot (e''=\mathit{Vul}\vee\mathit{Vul}[S_0]) \land 
e'=\mathit{Vul}\\
%RExt2
&&\hspace{1 mm}{}\land
[(\mathit{time}(S_0) = 0 \land e = e''\land\lnot \mathit{At}(I_1)[S_0] \land
\mathit{At}(I_0)[S_0]\\
&&\hspace{7.5 mm}{}\land (e'' = Vul \lor \lnot \mathit{Vul}[S_0]))\vee{}\\
&&\hspace{5.5 mm}(\mathit{time}(S_0) > 0 \land \phi_a[S_o]\\
&&\hspace{7.5 mm} {}\land \mathcal{R}_\mathit{ext}[\phi_a[do(move(I_0,I_1,e''),S_0)]]%\\ &&\hspace{2 mm}{}
\land 
\mathit{false})\vee{}\\
&&\hspace{5.5 mm}(\mathit{time}(S_0) > 0\wedge \lnot \phi_a \\
&&\hspace{7.5 mm} {}
\land \mathcal{R}_\mathit{ext}[\phi_a[do(move(I_0,I_1,e''),S_0)]] \land\mathit{false})]))\\
&&\\
%%%%%%
&&\hspace{-7 mm}{}\equiv\neg(e'=\mathit{Vul}\vee e''=\mathit{Vul}\vee\mathit{Vul}[S_0])\vee{}\\
&&\hspace{-7 mm}(\exists e.(0> -1 \land(e''=\mathit{Vul}\vee\mathit{Vul}[S_0]) \\
&&\hspace{1 mm}{}\land\mathit{true} \land e = e''\land\lnot \mathit{Vul}[S_0]\land e''=\mathit{Vul})\vee{}\\
&&\hspace{-1.5 mm}(0> -1\wedge\lnot (e''=\mathit{Vul}\vee\mathit{Vul}[S_0]) \land 
e'=\mathit{Vul}\\
%RExt2
&&\hspace{1 mm}{}\land
\mathit{true} \land e = e''\land 
\lnot \mathit{At}(I_1)[S_0] \land
\mathit{At}(I_0)[S_0]\\
&&\hspace{1 mm}{}\land (e'' = Vul \lor \lnot \mathit{Vul}[S_0]))\\
%\lnot \mathcal{R}_\mathit{ext}[\phi_a[S_0]]\land e''=\mathit{Vul}))
%\\
%&&\\
%%%%%%
% &&\hspace{-7 mm}{}\equiv\neg(e'=\mathit{Vul}\vee e''=\mathit{Vul}\vee\mathit{Vul}[S_0])\vee{}\\
% &&\hspace{-7 mm}(\exists e.(e''=\mathit{Vul}\land e = e''\land\lnot \mathit{Vul}[S_0])\vee{}\\
%
% &&\hspace{-1.5 mm}(\lnot e''=\mathit{Vul}\land\lnot\mathit{Vul}[S_0] \land 
% e'=\mathit{Vul}\\
% %RExt2
% &&\hspace{1 mm}{}\land
% e = e''\land\lnot \mathcal{R}_\mathit{ext}[\phi_a[S_0]]\land e''=\mathit{Vul}))\\
% &&\\
%%%%%%%
% &&\hspace{-7 mm}{}\equiv\neg(e'=\mathit{Vul}\vee e''=\mathit{Vul}\vee\mathit{Vul}[S_0])\vee{}\\
% &&\hspace{-7 mm}(\exists e.(e''=\mathit{Vul}\land e = e''\land\lnot \mathit{Vul}[S_0])\vee\mathit{false})\\
% &&\\
% %%%%%%%
% &&\hspace{-7 mm}{}\equiv(\neg(e'=\mathit{Vul}\vee e''=\mathit{Vul}\vee\mathit{Vul}[S_0])\vee{}\\
% &&\hspace{1 mm}(\exists e.\;e''=\mathit{Vul}\land e = e''\land\lnot \mathit{Vul}[S_0])).
\end{eqnarray*}
\end{small}
\par\noindent
By Axiom (6) on page 3, this is equivalent to:
\par\noindent
\begin{small}
\begin{eqnarray*}
&&\hspace{-7 mm}{}\equiv\{(e' \neq \mathit{Vul}\land e'' \neq \mathit{Vul})
\vee e''=\mathit{Vul} \lor 
(e'' \neq \mathit{Vul} \land e' = \mathit{Vul}) \}.
\end{eqnarray*}
\end{small}
\par\noindent
Collecting these results together, thus our third application of $\mathcal{R}_\mathit{ext}$ gives us the following:
\par\noindent
\begin{small}
\begin{eqnarray*}
&&\hspace{-7 mm}\mathcal{R}_\mathit{ext}[\mathit{CAfter}(\epsilon,\forall e''.\;\mathit{Poss}(\mathit{move}(I_0,I_1,e''))\\
&&\hspace{19 mm}{}\supset\forall e'.(\psi_a\supset\psi_b),S_0)]\\
&&\\
%%%
&&\hspace{-7 mm}{}=\forall e''.\;\mathcal{R}_\mathit{ext}[\mathit{Poss}(\mathit{move}(I_0,I_1,e''),S_0)]\\
&&\hspace{0 mm}{}\supset\forall e'.(\mathcal{R}_\mathit{ext}[\psi_a[S_0]]\supset\mathcal{R}_\mathit{ext}[\psi_b[S_0]])\\
&&\\
%%%
&&\hspace{-7 mm}{}\equiv\forall e''.\;\{e''=\mathit{Vul}\vee e''=\mathit{NotVul}\}\\
&&\hspace{5 mm}{}\supset\forall e'.\;(\{e'=\mathit{Vul}\vee e'=\mathit{NotVul}\}\\
&&\hspace{10 mm}{}\supset \{(e' \neq \mathit{Vul}\land e'' \neq \mathit{Vul})
\vee e''=\mathit{Vul}\\
&&\hspace{16 mm}{}\lor 
(e'' \neq \mathit{Vul} \land e' = \mathit{Vul}) \}\\
&&\\
&&\hspace{-7 mm}{}\equiv true.\hspace{55 mm}\Box
\end{eqnarray*}
\end{small}

\begin{comment}
%ZZZ
\noindent\textbf{\color{red}Work in progress below!}\\\noindent\textbf{\color{red}Work in progress below!}
\begin{small}
\begin{eqnarray*}
&&\hspace{-7 mm}\mathcal{R}_\mathit{ext}[\mathit{CAfter}(\epsilon,\forall e''.\;\mathit{Poss}(\mathit{move}(I_0,I_1,e''))\supset\psi',S_0)]={}\\
%
&&\hspace{-7 mm}\forall e''.\;e''=\mathit{Vul}\vee e''=\mathit{NotVul}\supset{}\\
&&\hspace{-4 mm}\{\neg(e'=\mathit{Vul}\vee e''=\mathit{Vul}\vee\mathit{Vul}[S_0])\vee{}\\
&&\hspace{-2 mm}(\exists e.\;\mathit{time}(S_0)= 0 \land e''=\mathit{Vul}\land e = e''\land\lnot \mathit{Vul}[S_0])\}.
\end{eqnarray*}
\end{small}
\end{comment}
% This trivially follows.
% {\color{red}\textbf{[NOPE!!!]}}.
%\\
%\\
%{\color{red}\textbf{[Made a mistake in the 2nd application!!]}}

\end{document}